\documentclass[10pt,twocolumn,letterpaper]{article}

\usepackage[pagenumbers]{cvpr} 

\definecolor{cvprblue}{rgb}{0.21,0.49,0.74}
\usepackage[pagebackref,breaklinks,colorlinks,allcolors=cvprblue]{hyperref}

\usepackage{amsmath,amssymb,amsthm}
\usepackage{bm}
\usepackage{graphicx}
\usepackage{hyperref}
\usepackage{enumitem}
\usepackage{algorithm}
\usepackage{xpatch}
\usepackage{algpseudocode}
\usepackage{multirow}
\usepackage[table]{xcolor}
\usepackage{pifont}
\usepackage{tcolorbox}
\usepackage{booktabs}       
\usepackage{amsfonts}       
\usepackage{nicefrac}       
\usepackage{microtype}      
\usepackage{colortbl} 
\usepackage{tikz}
\DeclareMathOperator*{\argmin}{arg\,min}

\def\paperID{674} 
\def\confName{CVPR}
\def\confYear{2026}

\title{\textit{UniComp}: Rethinking Video Compression Through Informational Uniqueness}

\author{
Chao Yuan$^{1,2,\S}$, Shimin Chen$^{1}$, Minliang Lin$^{1}$, Limeng Qiao$^{1}$, Guanglu Wan$^{1}$, Lin Ma$^{1,\dagger,\S}$ \\
\textsuperscript{1} Meituan Inc. \textsuperscript{2} Beihang University
\\
{\tt\small yuanc3666@gmail.com, shiminchen1996@163.com, forest.linma@gmail.com} \\ 
}

\begin{document}

\twocolumn[{
\maketitle
\begin{center}
    \centering
    \includegraphics[width=0.87\textwidth]{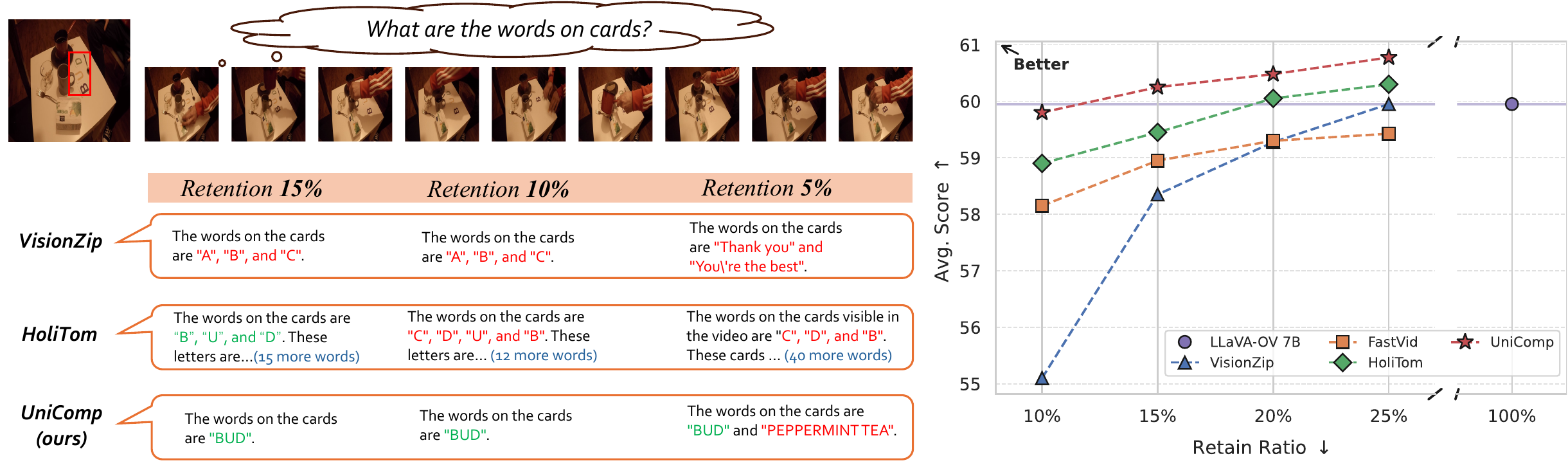} 
    \captionof{figure}{\textbf{Left:} Compare UniComp with state-of-the-art methods (VisionZip~\cite{yang2025visionzip} and HoliTom~\cite{shao2025holitom}) on Eagle2.5~\cite{chen2025eagle} model under three retained ratio settings. Input is 32 frames with 256 tokens of each frame. Words in \textcolor{green}{Green} means right, and \textcolor{red}{Red} means wrong. UniComp could recognize even only retained 5\% tokens, although contains wrong words, but the words ``PEPPERMINT TEA" on the tea box, which is surprising. \textbf{Right:} Performance compared to SOTA methods, UniComp could even surpass baseline which has not been compressed.}
    \label{fig:teaser}
\end{center}%
}]
\renewcommand{\thefootnote}{}
\footnotetext{$\dagger$ Corresponding author, $\S$ During his work at Meituan.}
\footnotetext{Codes: \textit{\url{https://github.com/TimeMarker-LLM/UniComp}}}

\begin{abstract}

Distinct from attention-based compression methods, this paper presents an \textbf{information \underline{uni}queness driven video \underline{comp}ression framework}, termed \textbf{UniComp}, which aims to maximize the information fidelity of video representations under constrained computational budgets. Starting from the information-theoretic perspective, we formulate the vision compression as an optimization problem that minimizes conditional entropy (reconstruction error) between retained and full tokens. To achieve this, we introduce the notion of \textbf{information uniqueness} to measure intrinsic redundancy among tokens to link with reconstruction error. Based on uniqueness, we design three modules—\textbf{Frame Group Fusion}, \textbf{Token Allocation}, and \textbf{Spatial Dynamic Compression}—that progressively perform semantic frame grouping, adaptive resource allocation, and fine-grained spatial compression. Extensive experiments demonstrate that \textbf{UniComp} consistently outperforms existing compression methods in preserving essential visual tokens under limited computational budgets, highlighting the pivotal role of information uniqueness in token compression efficacy. 

\end{abstract}

\section{Introduction}
\label{sec:intro}

With the rapid advancement of multimodal large models (MLLMs), the computational cost of processing dense video inputs has become a critical bottleneck for scalability and efficiency. Recent works~\cite{yang2025visionzip,tao2025dycoke,huang2024prunevid,shen2025fastvid,shao2025holitom} have made notable progress in modeling feature redundancy and compressing spatial-temporal information, offering valuable inspiration for efficient video understanding. However, most existing methods primarily rely on attention-based importance scoring. While effective in highlighting salient content, these methods often introduce redundancy across frames and tokens and tend to overlook fine-grained details. Under aggressive compression settings, it toward redundancy saliency can lead to loss of essential information. Furthermore, state-of-the-art methods like FastVid~\cite{shen2025fastvid} and HoliTom~\cite{shao2025holitom} require tuning more than five hyper-parameters, and inner-LLM modification approaches like DyCoke~\cite{tao2025dycoke} and HoliTom~\cite{shao2025holitom} involve altering attention layers—making them difficult to generalize across architectures.

\begin{figure}[t]
\centering
\includegraphics[width=\linewidth]{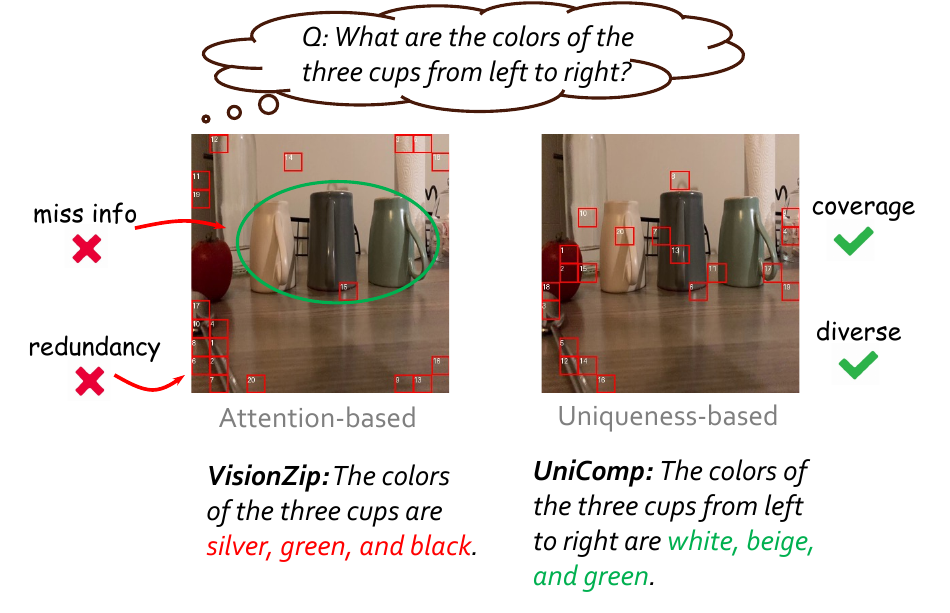}
\caption{Real visualization on LLaVA-OneVision-7B~\cite{li2024llava} of differences between attention-based (like VisionZip~\cite{yang2025visionzip} and HoliTom~\cite{shao2025holitom}) and our uniqueness-based token selection. We select top-20 tokens with red rectangles and labeled with orders. Baseline attention-based selection is redundant and misses key content, while ours captures essential information with diverse coverage.} 
\label{fig:intro}
\end{figure}

In this work, we advocate a new perspective: \textbf{the essence of video compression lies not in `attention' but in `information uniqueness'.} As shown in Fig.~\ref{fig:intro}, we argue that, under constrained computational budgets, the model should prioritize retaining frames and tokens that carry unique and irreplaceable information. Because redundant or overlapping representations can be compacted or reconstructed from others. This perspective provides an information-theoretic foundation for understanding compression fidelity.


Formally, we model the compression process as minimizing the conditional entropy $H(\mathcal{X} \mid \mathcal{S})$ between selected tokens $\mathcal{S}$ and the full token set $\mathcal{X}$, which is equivalent to minimizing reconstruction error. Then, we derive an upper bound \textbf{linking information uniqueness to reconstruction error}, thus establishing a theoretical bridge between uniqueness and information compression.

Base on this principle, we propose \textbf{\textit{UniComp}}, an information uniqueness-driven video compression framework composed of three synergistic modules:
\ding{192} \textbf{Frame Group Fusion (FGF)} - adaptively merges temporally redundant frames based on semantic uniqueness to reduce temporal redundancy;
\ding{193} \textbf{Token Allocation (TA)} - allocates token budgets according to global frame uniqueness, dynamically distributing computation toward more informative content;
\ding{194} \textbf{Spatial Dynamic Compression (SDC)} greedily selects and fuses tokens within each frame based on token-level uniqueness, further eliminating local redundancy. 
Unified under the principle of maximizing information uniqueness, \textbf{\textit{UniComp}} achieves efficient and information-preserving compression across temporal, spatial, and global dimensions. 

\textit{UniComp} is plug-and-play, requiring only two hyper-parameters, with default settings transferable across different ViTs and LLMs. Unlike inner-LLM modification methods, \textit{UniComp} can be applied to other models with minimal code changes, making it highly generalizable. As shown in Fig.~\ref{fig:teaser}, extensive experiments demonstrate that \textit{UniComp} consistently outperforms existing compression paradigms under limited computational budgets.

In summary, our key contributions are as follows:


\begin{itemize}
    \item We provide an information-theoretic formulation of information compression by minimizing conditional entropy, and introduce the notion of \textbf{information uniqueness} to quantify feature redundancy. This establishes a theoretical link between uniqueness and information fidelity.
    \item We propose \textbf{\textit{UniComp}}, a uniqueness-driven compression framework that integrates temporal fusion, global allocation, and spatial compression under a unified principle-keep unique.
    \item High generalizability with few hyper-parameters and few implementation overhead, enabling plug-and-play deployment across architectures.
    \item State-of-the-art performance on multiple long video understanding benchmarks, consistently achieving superior compression efficiency and semantic fidelity under varying computational budgets.
\end{itemize}

\setlength{\parindent}{1em}
\section{Related Work}

\subsection{Video Large Language Models}
Recent advances in multimodal large language models (MLLMs)~\cite{li2024llava,chen2025eagle,xiaomi2025mimo,bai2025qwen2,guo2025seed1,liu2025nvila,shen2025longvita} have significantly expanded the capabilities of video understanding~\cite{maaz2024video,team2025kwai,chen2024timemarker,yuan2025date,shu2025audio,wu2025number,liu2025vidcom2}. Early approaches~\cite{chen2023videollm,he2024vlab,huang2024vtimellm,yang2022frozenbilm,chen2024fewer,lin2024video,li2024llava} typically process raw video tokens directly or employ simple pooling strategies, while improving video performance through task transfer from image-based models. Subsequently, models like LLaVA-Video~\cite{zhang2024video}, BT-Adapter~\cite{liu2023one}, and VILA~\cite{lin2024vila} enhance spatio-temporal modeling. More recently, a series of Video LLMs incorporating training-time compression have emerged, including MovieChat~\cite{song2024moviechat}, LLaMA-VID~\cite{li2024llamavid}, Chat-UniVi~\cite{jin2024chat}, LongVU~\cite{shen2024longvu}, and Minicpm-v 4.5~\cite{yu2025minicpm}. These methods reduce redundancy via techniques such as adaptive pooling, token clustering, or 3D resampling, thereby enabling longer video sequences. However, they typically require fine-tuning and substantial computational resources, which limits the practicality for deployment across diverse architectures.
 
\subsection{Token Compression in Vision Language Models}
Visual token compression has become an essential strategy for mitigating redundancy in MLLMs. Spatial compression approaches such as ToMe~\cite{bolya2022token} aggregate semantically similar tokens in vision transformers (ViTs) to reduce spatial redundancy. Other methods predominantly rely on attention-score analysis; for example, VisionZip~\cite{yang2025visionzip} and SparseVLM~\cite{zhang2024sparsevlm} leverage attention scores in ViTs to eliminate spatial redundancy. In addition, FastV~\cite{chen2024image} prunes non-critical visual tokens in shallow layers of LLMs, and PDrop~\cite{xing2024pyramiddrop} performs progressive token pruning across multiple stages within LLM. However, they often overlook temporal dependencies, thereby limiting effectiveness in video scenarios.

Specialized video compression methods address spatio-temporal redundancy. DyCoke~\cite{tao2025dycoke} merges tokens across frames and reduces KV cache size; PruneVID~\cite{huang2024prunevid} clusters tokens based on query relevance; FastVID~\cite{shen2025fastvid} combines temporal segmentation on frame sequences; FrameFusion~\cite{fu2024framefusion} and Holitom~\cite{shao2025holitom} integrates token merging across LLM layers. While effective, many rely on heuristic importance estimation, which may degrade performance under extreme compression. The challenges of maintaining semantic fidelity under extreme compression and difficulty of transferring model-specific solutions motivate \textit{UniComp}.

\section{Methods}

\subsection{Information Uniqueness and Optimization Objective}
\label{value}
In contrast to attention-based compression approaches, \textbf{\textit{UniComp}} operates under the principle of \textbf{\textit{information uniqueness}}. We prioritize retaining frames or tokens with high uniqueness, while redundant ones are merged or removed, whose information is largely contained in other frames/tokens and can be inferred or reconstructed from the retained representations.

\paragraph{Information Uniqueness.}
We define the pairwise \emph{\textbf{uniqueness}} between token $x_i$ and $x_j$ as:
\begin{equation}
u_{ij} = 1-\frac{x_i^\top x_j}{|x_i| |x_j|}
\label{uniq1}
\end{equation}
where $|\cdot|$ is the Euclidean norm.

The \textbf{uniqueness} $U_i$ of token $i$ is then defined as the average uniqueness between $x_i$ and all $N$ tokens:
\begin{equation}
U_i = 1-\frac{1}{N}\sum_{j=1}^{N}{\frac{x_i^\top x_j}{|x_i| |x_j|}}
\label{uniq2}
\end{equation}

\paragraph{Optimization Objective.}
Our core hypothesis is that, given a compression budget (i.e., retained tokens $K$), the selected subset $\mathcal{S}$ should preserve maximal information from the full token set $\mathcal{X}=\{x_1,\dots,x_N\}$ and a budget $|\mathcal{S}|=K$, we define the optimization objective:, such that the information of discarded tokens can be reconstructed or approximated from $\mathcal{S}$. Formally, this is equivalent to minimizing the \textbf{information loss} introduced by compression, or, from an information-theoretic perspective, maximizing the information uniqueness of the retained tokens.


We start from the conditional entropy formulation $H(\mathcal{X}\mid\mathcal{S})$ to \textbf{minimize information loss}:
\begin{align}
\mathcal{S}^* &= \argmin\limits_{S \subset X, \, |S| =K} H(\mathcal{X}\mid\mathcal{S})
\end{align}
where $H(\mathcal{X}\mid\mathcal{S})$ measures the information loss when only $\mathcal{S}$ is retained.

\paragraph{From Information Loss to Reconstruction Error.} 
Assuming the token features follow an isotropic Gaussian distribution, minimizing $H(\mathcal{X}\mid\mathcal{S})$ is equivalent to minimizing the reconstruction error variance.
\begin{equation}
\begin{aligned}
    &H(\mathcal{X}\mid\mathcal{S}) \propto \log \det \Sigma_{\mathcal{X}\mid\mathcal{S}}  \Rightarrow \\
    &\arg\min H(\mathcal{X}\mid\mathcal{S}) = \arg\min \mathcal{E}(\mathcal{S})
\end{aligned}
\end{equation}
where the reconstruction error $\mathcal{E}(\mathcal{S})$ is defined as:
\begin{equation}
\mathcal{E}(\mathcal{S}) = \sum_{j\in \mathcal{X}} \|x_j - \hat x_j\|^2
\end{equation}
where $\hat x_j$ is the reconstruction of $x_j$ using tokens in $\mathcal{S}.$



\begin{figure}[t]
\centering
\includegraphics[width=0.5\textwidth]{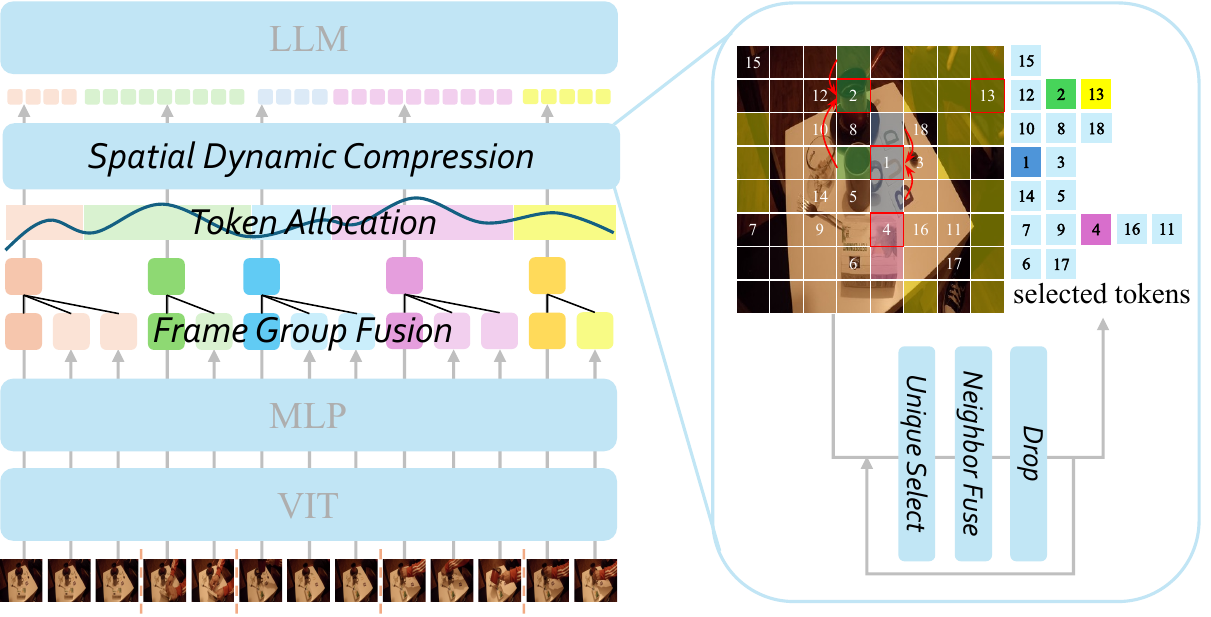}
\caption{Framework of \textit{UniComp}. It has three modules: Frame Group Fusion (FGF), Token Allocation (TA), and Spatial Dynamic Compression (SDC). Right part shows retained tokens selection and fusion. Red rectangles are retained token labeled with orders, and we visualize four token (1/2/4/13) fusion with four colors (token with the same color will be fused into the token with red rectangle).}
\label{fig:framework}
\end{figure}


\paragraph{Uniqueness-Based Upper Bound.} We further derive (see \textbf{Supplementary A} for the proof) that the reconstruction error admits the following upper bound related to our defined \textbf{uniqueness} (Eq.~\ref{uniq1}):

\begin{equation}
\mathcal{E}(\mathcal{S}) \leq 2 \sum_{j\in\mathcal{X}} \min_{i\in\mathcal{S}}u_{ij} 
\label{optm}
\end{equation}






This bound explicitly \textbf{links reconstruction error to uniqueness}: if the discarded tokens have high uniqueness, the reconstruction error increases.

Therefore, minimizing this upper bound naturally leads to a greedy selection strategy for $S$ based on information uniqueness, which forms the core of \textit{UniComp}.

\subsection{Frame Group Fusion}
\label{frams_fuse}

Video sequences often contain substantial temporal redundancy, as consecutive frames tend to share highly similar semantics. To preserve video information while reducing redundant temporal information, we perform an adaptive \textbf{Frame Group Fusion} (FGF) mechanism guided by frame \textbf{uniqueness},  in Fig.~\ref{fig:framework}. This module aims to fuse highly redundant frames into compact representative groups.

Given a sequence of video frames embeddings $\mathcal{F} = \{f_1, f_2, \dots, f_T\}$ which obtained from the ViT encoder, we first obtain a \textbf{global frame feature} for each frame by average pooling all visual tokens:
\begin{equation}
\small
\hat{f}_t = \frac{1}{M} \sum_{m=1}^{M} f_{t,m}, \quad f_t \in \mathbb{R}^d
\end{equation}
where $f_{t,m}$ denotes the $m$-th token feature of frame $F_t$, and $M$ is the number of tokens per frame. This is a faster and costless way, and alternatively, using the classify head of ViT to get a global feature is more accurate but slightly slower.


Base on the definition of uniqueness (Eq. \ref{uniq1}), we then define uniqueness score $u(f_i, f_j)$ to measure the semantic difference between frames $i$ and $j$. Based on this, the video sequence is scanned sequentially to form frame groups.

Starting from the first frame $f_1$, we initialize a new group $G_1 = {f_1}$.
For each subsequent frame $f_t$, if its uniqueness difference from the first representative frame $f_r$ (the first frame of the group, e.g. $r=1$ for $G_1$) of the current group satisfies $u(f_t, f_r) < U_f$ ($U_f$ is a hyper-parameter), the frame $t$ is considered semantically redundant and merged into the same group. Otherwise, a new semantic segment emerges, and a new group is created as $G_{k+1} = {x_t}$.

Each group is then fused into a single representative feature via mean pooling:
\begin{equation}
\small
\tilde{f}_k = \frac{1}{|G_k|} \sum_{f_t \in G_k} f_t
\end{equation}

The resulting sequence of representative features $\tilde{\mathcal{F}} = \{\tilde{f}_1, \tilde{f}_2, \dots, \tilde{f}_{K_f}\}$ serves as the input for subsequent token allocation and compression modules, where $K_f$ is the number of groups, enabling structured temporal compression with maximal information preservation.

This uniqueness-driven grouping strategy adaptively adjusts the temporal information according to semantic variation:
in stable scenes or low-motion regions, multiple frames are merged to suppress redundancy; while in segments with large semantic transitions, grouping becomes finer to retain key dynamic information.

\subsection{Token Allocation}

After obtaining the fused frame representations, we further perform adaptive \textbf{Token Allocation} (TA) according to the global uniqueness of each frame.
The intuition is that \textbf{frames with higher uniqueness} (reflecting larger semantic deviations from others) \textbf{should be allocated more visual tokens to preserve critical information}, while \textbf{frames with lower uniqueness} receive fewer tokens, since they can get potential information from other frame for compensation since their features are more similar to others.

Formally, given the global features obtained in Sec.~\ref{frams_fuse} $\tilde{\mathcal{F}} = {\tilde{f}_1, \tilde{f}_2, \dots, \tilde{f}_{K_f}}$, we first compute the uniqueness of frame $t$ base on Eq.~\ref{uniq2}:
\begin{equation}
\small
U_t = 1-\frac{1}{K_f}\sum_{s=1}^{K_f}{\frac{\tilde{f}_t^\top \tilde{f}_s}{|\tilde{f}_t| |\tilde{f}_s|}}
\end{equation}

Then, we normalize frame uniqueness values with the mean. Since uniqueness is related to the sum of frames, the difference between uniqueness values will be greatly reduced as frames increases. Therefore, we scale it based on frames number $K_f$ to amplify the difference.
\begin{equation}
U_t = (U_t - \bar{U}) \cdot \sqrt{K_f}
\end{equation}
where $\bar{U}$ is the mean uniqueness among frames, and the scaling term $\sqrt{K_f}$ stabilizes distribution variance across different frame lengths.

Finally, a softmax transforms uniqueness scores into a probability distribution, which determines the proportional token allocation $K_t$ for each frame $t$:
\begin{equation}
\small
K_t = \left\lfloor \frac{e^{U_t}}{\sum_{s\in [1,K_f]} e^{U_s}} \cdot \text{TOKEN}_{\text{max}} \right\rfloor
\end{equation}
where $\text{TOKEN}_{\text{max}}$ denotes the total token budget across the sequence, and $K_t$ represents the number of tokens assigned to the $t$-th frame.

This allocation strategy dynamically adjusts computational focus according to semantic uniqueness: frames that contribute more unique visual content receive richer token representations, while those offering redundant information are compactly represented.


\subsection{Spatial Dynamic Compression}

After determining the token budget for each frame, as detailed in Fig.~\ref{fig:framework}, we further perform frame-wise \textbf{spatial dynamic compression} (SDC) based on uniqueness to eliminate redundancy and preserve the most informative and representative tokens. As discussed in Sec.~\ref{value}, this step corresponds to maximizing the information uniqueness within each frame, thereby realizing the optimization of maximizing information preservation.

\paragraph{(1) Token uniqueness.}
For the $t$-th frame, according to Eq.~(\ref{uniq1},\ref{uniq2}), the token uniqueness of the $i$-th token is:
\begin{equation}
U_i^{(t)} = \frac{1}{N}\sum_{j=1}^{N}u_{ij}^{(t)}
\end{equation}
where $u_{ij}^{(t)}$ is the uniqueness graph of frame $t$, and $N$ is the tokens of a frame (LLaVA-OV~\cite{li2024llava} is 196, LLaVA-Vid~\cite{zhang2024video} is 169, and Eagle2.5~\cite{chen2025eagle} is 256). A higher $U_i^{(t)}$ indicates that token $x_i^{(t)}$ is more distinctive and carries more irreplaceable information within the frame. Following \cite{yang2025visionzip}, we use the last attention layer's Keys of ViT to calculate the uniqueness, since Keys summarize
information contained in each token and have fewer feature dimensions for better representation.

\paragraph{(2) Retained token selection.}
As declared in Sec.~\ref{value}, to minimize the information loss is to minimize Eq.~\ref{optm}, which is related to the token uniqueness, so we designed a greedy strategy about uniqueness to achieve it.

We sort all tokens by their uniqueness scores in descending order and iteratively select representative tokens.
Given a constant uniqueness threshold $U_c$, if two tokens $i,j$ of frame $t$ satisfy $u^{(t)}_{i,j} < U_c$ they are considered redundant, which also means that MLLMs can obtain relevant information from other retained tokens as compensation to reconstruct the entire image content. Meanwhile, Instead of discarding them directly, we perform a \textbf{neighbor token fusion} step:
\begin{equation}
x_i^{(t)} \leftarrow
\frac{1}{2} \Big( x_i^{(t)} +
\frac{1}{|\mathcal{N}_i^{(t)}|}\sum_{j\in\mathcal{N}_i^{(t)}} x_j^{(t)} \Big)
\end{equation}
where $\mathcal{N}_i^{(t)} = \{j \mid u^{(t)}_{i,j} < U_c\}$ denotes the neighborhood of token $i$ filtered with the uniqueness graph.
This process can be viewed as a greedy graph redundancy elimination mechanism, where each step compresses a uniqueness cluster into a single representative token, effectively reducing token overlap.

The overall selection process can be summarized as:
\begin{algorithm}[H]
\small
\caption{Spatial Dynamic Compression}
\label{alg:SDC}
\begin{algorithmic}[1]
\Require For frame $t$, uniqueness matrix $u_{ij}^{(t)}$, token budget $K_t$, and constant uniqueness threshold $U_c$,
\Ensure Compressed frame tokens $\tilde{F}_t$
\State Compute global uniqueness: $U_i^{(t)} = \frac{1}{N}\sum_j u_{i,j}^{(t)}$
\State $\text{sort\_idx} \gets \operatorname{argsort}(U^{(t)}, \text{desc})$
\State Initialize selected ids set $\mathcal{S}_t \gets \emptyset$
\State Initialize Redundant ids set $\mathcal{R}_t \gets \emptyset$
\For{each $i$ in $\text{sort\_ids}$}
  \If{$i \in \mathcal{S}_t \cup \mathcal{R}_t$}
    \State \textbf{continue}
  \EndIf
  \State Add $i$ to $\mathcal{S}_t$
  \State $\mathcal{N}_i^{(t)} \gets \{ j \mid u^{(t)}_{i,j} < U_c \}$
  \State Fuse: $x_i^{(t)} \gets \frac{1}{2}(x_i^{(t)} + \frac{1}{|\mathcal{N}_i^{(t)}|}\sum_{j\in\mathcal{N}_i^{(t)}}x_j^{(t)})$
  \State Add $\mathcal{N}_i^{(t)}$ to $\mathcal{R}_t$
  \If{$|\mathcal{S}_t| \ge K_t$}
    \State \textbf{break}
  \EndIf
\EndFor
\State $\tilde{F}_t \gets [x_i^{(t)} \mid i \in \mathcal{S}_t]$
\State \Return $\tilde{F}_t$
\end{algorithmic}
\end{algorithm}

Considering that \textbf{this process involves causal loops} and \textbf{exhibits extremely high time complexity}, we optimized its implementation \textbf{using matrix-level parallel computation}, reducing the overall computational complexity by nearly \textbf{20× without any performance degradation}.

The order of selected retained tokens here is the original token order, not sorted by the uniqueness order, although their performance differences on benchmarks are minor, as discussed in the experimental section.

Furthermore, considering the possibility of $K_t$ exceeding the frame token limit and causing waste, we perform a re-allocation of wasted tokens to maximize the utilization of maximum token limit $\text{TOKEN}_{\text{max}}$. Specifically, we allocate the wasted tokens evenly among frames that haven't exceeded the limit. Additionally, due to our dynamic token setting, the original model struggles to distinguish the boundaries of tokens between frames. Therefore, we add a newline token after each frame to isolate inter-frame information. This token is also included in the maximum token limit.


\begin{table*}[t]
\centering
\small
\setlength{\tabcolsep}{4pt}
\renewcommand{\arraystretch}{0.8}
\caption{Comparison of state-of-the-art video compression methods across benchmarks with \textbf{long videos}. We test under 4 different retained ratios (25\%, 20\%, 15\%, 10\%) with 32 frames input on LLava-OneVision-7B model. UniComp consistently achieves the best average performance under various frame-retention ratios. \textbf{Bold} is the best method and \underline{underline} is the second. Acc. is the ratio to baseline.}
\begin{tabular}{lc|cccc|cc}
\toprule
\textbf{Method} & \textbf{Frames/Retain Ratio} & \textbf{LongVideoBench} & \textbf{EgoSchema} & \textbf{MLVU} & \textbf{VideoMME} & \textbf{Avg.} & \textbf{Acc (\%)} \\
\midrule
\rowcolor{gray!15}
Vanilla & 32f / 100\% & 56.3 & 60.4 & 64.7 & 58.4 & 59.95 & 100.0 \\
\midrule
FastV~\cite{chen2024image} $_{\text{ECCV'24}}$& 32f / 25\% & 56.4 & 60.4 & 61.5 & 56.1 & 58.60 & 97.7 \\
Pdrop~\cite{xing2024pyramiddrop} $_{\text{CVPR'25}}$& 32f / 25\% & \underline{56.7} & 58.0 & - & 56.4 & - & - \\
DyCoke~\cite{tao2025dycoke} $_{\text{CVPR'25}}$& 32f / 25\% & 49.5 & 59.9 & 55.8 & 57.4 & 55.65 & 92.8 \\
VisionZip~\cite{yang2025visionzip} $_{\text{CVPR'25}}$ & 32f / 25\% & 56.5 & 60.3 & \underline{64.8 }& 58.2 & 59.95 & 100.0 \\
PruneVid~\cite{huang2024prunevid} $_{\text{ACL'25}}$ & 32f / 25\% & 55.7 & 59.9 & - & 57.4 & - & - \\
FastVid~\cite{shen2025fastvid} $_{\text{NeurIPS'25}}$& 32f / 25\% & 56.3 & 59.3 & 64.1 & 58.0 & 59.43 & 99.1 \\
HoliTom~\cite{shao2025holitom} $_{\text{NeurIPS'25}}$ & 32f / 25\% & \underline{56.7} & \underline{61.2} & 64.7 & \underline{58.6} & \underline{60.30} & \underline{100.6} \\
\rowcolor[HTML]{EAFAF1} 
\textbf{UniComp} & 32f / 25\% & \textbf{57.6} & \textbf{61.6} & \textbf{65.0} & \textbf{58.9} & \textbf{60.78} & \textbf{101.4} \\
\midrule
VisionZip~\cite{yang2025visionzip} $_{\text{CVPR'25}}$ & 32f / 20\% & 55.2 & 59.8 & 64.2 & 57.9 & 59.28 & 98.9 \\
PruneVid~\cite{huang2024prunevid} $_{\text{ACL'25}}$& 32f / 20\% & 54.7 & 59.7 & - & 56.9 & - & - \\
FastVid~\cite{shen2025fastvid} $_{\text{NeurIPS'25}}$& 32f / 20\% & \underline{57.1} & 58.3 & \underline{63.9} & 57.9 & 59.30 & 98.9 \\
HoliTom~\cite{shao2025holitom} $_{\text{NeurIPS'25}}$ & 32f / 20\% & \underline{57.1} & \underline{61.0} & 63.5 & \underline{58.6} & \underline{60.05} & \underline{100.2} \\
\rowcolor[HTML]{EAFAF1} 
\textbf{UniComp} & 32f / 20\% & \textbf{57.7} & \textbf{61.1} & \textbf{64.4} & \textbf{58.7} & \textbf{60.48} & \textbf{100.9} \\
\midrule
VisionZip~\cite{yang2025visionzip} $_{\text{CVPR'25}}$ & 32f / 15\% & 54.4 & 59.8 & 63.1 & 56.1 & 58.35 & 97.3 \\
PruneVid~\cite{huang2024prunevid} $_{\text{ACL'25}}$& 32f / 15\% & 55.4 & 59.7 & - & 56.6 & - & - \\
FastVid~\cite{shen2025fastvid} $_{\text{NeurIPS'25}}$& 32f / 15\% & 56.2 & 58.7 & \underline{63.2} & \underline{57.7} & 58.95 & 98.3 \\
HoliTom~\cite{shao2025holitom} $_{\text{NeurIPS'25}}$ & 32f / 15\% & \underline{56.4} & \textbf{61.2} & 62.9 & 57.3 & \underline{59.45} & \underline{99.2} \\
\rowcolor[HTML]{EAFAF1} 
\textbf{UniComp} & 32f / 15\% & \textbf{57.4} & \underline{60.9} & \textbf{64.3} & \textbf{58.4} & \textbf{60.25} & \textbf{100.5} \\
\midrule
VisionZip~\cite{yang2025visionzip} $_{\text{CVPR'25}}$ & 32f / 10\% & 49.3 & 58.0 & 59.7 & 53.4 & 55.10 & 91.9 \\
PruneVid~\cite{huang2024prunevid} $_{\text{ACL'25}}$& 32f / 10\% & 54.5 & 59.8 & 62.3 & 56.0 & 58.15 & 97.0 \\
FastVid~\cite{shen2025fastvid} $_{\text{NeurIPS'25}}$& 32f / 10\% & \underline{56.3} & 58.3 & \underline{62.7} & \underline{57.3} & 58.65 & 97.8 \\
HoliTom~\cite{shao2025holitom} $_{\text{NeurIPS'25}}$ & 32f / 10\% & \underline{56.3} & \textbf{61.2} & 61.3 & 56.8 & \underline{58.90} & \underline{98.2} \\
\rowcolor[HTML]{EAFAF1} 
\textbf{UniComp} & 32f / 10\% & \textbf{57.6} & \underline{60.5} & \textbf{63.1} & \textbf{58.0} & \textbf{59.80} & \textbf{99.7} \\
\midrule
\bottomrule
\end{tabular}
\label{tab:results1}
\end{table*}

\begin{table*}[t]
\centering
\small
\setlength{\tabcolsep}{4pt}
\renewcommand{\arraystretch}{0.8}
\caption{The video compression performance under \textbf{large number of frames} input. Since LLaVA-onevison's official evaluation limitation is 32 frames (total 32*196=6272 tokens), so we sample 128, 256 and 320 frames and compress them to 32 frames limitation (6272 tokens).}
\begin{tabular}{lcc|cccc|cc}
\toprule
\textbf{Method} & \textbf{Frames/Retain Ratio} & \textbf{Max Tokens}&\textbf{LongVideoBench} & \textbf{EgoSchema} & \textbf{MLVU} & \textbf{VideoMME} & \textbf{Avg.} & \textbf{Acc (\%)} \\
\midrule
\rowcolor{gray!15}
Vanilla & 32f / 100\% &  6272 & 56.3 & 60.4 & 64.7 & 58.4 & 59.95 & 100.0 \\
\midrule
VisionZip~\cite{yang2025visionzip} $_{\text{CVPR'25}}$& 128f / 25\% &  6272 &57.8 & 61.1 & 68.7 & 59.7 & 61.83 & 103.1 \\
FastVid~\cite{shen2025fastvid} $_{\text{NeurIPS'25}}$& 128f / 25\% &  6272 &55.7 & 59.3 & 67.4 & 57.5 & 59.98 & 100.0 \\
HoliTom~\cite{shao2025holitom} $_{\text{NeurIPS'25}}$& 128f / 25\% &  6272 &58.6 & 61.5 & 68.1 & 60.7 & 62.23 & 103.8 \\
\rowcolor[HTML]{EAFAF1} 
\textbf{UniComp} & 128f / 25\% &  6272 &\textbf{59.0} & \textbf{62.4} & \textbf{68.6} & \textbf{59.7} & \textbf{62.43} & \textbf{104.1} \\
\midrule
VisionZip~\cite{yang2025visionzip} $_{\text{CVPR'25}}$& 256f / 12.5\% &  6272 &54.7 & 61.7 & 68.1 & 58.4 & 60.73 & 101.3 \\
FastVid~\cite{shen2025fastvid} $_{\text{NeurIPS'25}}$& 256f / 12.5\% &  6272 &52.3 & 57.3 & 62.6 & 55.9 & 57.03 & 95.1 \\
HoliTom~\cite{shao2025holitom} $_{\text{NeurIPS'25}}$& 256f / 12.5\% &  6272 &56.2 & 61.2 & 69.1 & 60.0 & 61.63 & 102.8 \\
\rowcolor[HTML]{EAFAF1} 
\textbf{UniComp} & 256f / 12.5\% &  6272 &\textbf{57.4} & \textbf{62.4} & \textbf{69.4} & \textbf{60.5} & \textbf{62.43} & \textbf{104.1} \\
\midrule
VisionZip~\cite{yang2025visionzip} $_{\text{CVPR'25}}$& 320f / 10\% &  6272 &53.4 & 60.7 & 66.7 & 57.3 & 59.53 & 99.3 \\
FastVid~\cite{shen2025fastvid} $_{\text{NeurIPS'25}}$& 320f / 10\% &  6272 &50.6 & 55.9 & 61.4 & 54.7 & 55.65 & 92.8 \\
HoliTom~\cite{shao2025holitom} $_{\text{NeurIPS'25}}$& 320f / 10\% &  6272 &56.5 & 61.1 & 68.2 & 59.9 & 61.43 & 102.5 \\
\rowcolor[HTML]{EAFAF1} 
\textbf{UniComp} & 320f / 10\% &  6272 &\textbf{58.5} & \textbf{61.2} & \textbf{69.4} & \textbf{60.7} & \textbf{62.45} & \textbf{104.2} \\
\midrule
\bottomrule
\end{tabular}
\label{tab:compression_results2}
\end{table*}

\begin{table}[t]
\centering
\small
\setlength{\tabcolsep}{3pt}
\renewcommand{\arraystretch}{0.9}
\caption{Comparison under different frame and compression settings on \textbf{LLaVA-Video-7B}~\cite{zhang2024video} under LongVideoBench (LVB) and MLVU benchmarks. “OOM” means out-of-memory. \textit{UniComp} consistently outperforms baselines. F/R is Frames/Ratain ratio.}
\begin{tabular}{lccc|ccc}
\toprule
\multirow{2}{*}{\textbf{Method}} & 
\multicolumn{3}{c}{\textbf{Short Setting}} & 
\multicolumn{3}{c}{\textbf{Long Setting}} \\
\cmidrule(lr){2-4} \cmidrule(lr){5-7}
 & \textbf{F/R} & \textbf{LVB} & \textbf{MLVU} & \textbf{F/R} & \textbf{LVB} & \textbf{MLVU} \\
\midrule
VisionZip & 64f/10\% & 52.2 & 62.7 & 256f/25\% & 57.2 & 70.1 \\
HoliTom & 64f/10\% & 54.8 & 63.3 & 256f/25\% & 57.3 & 69.8 \\
\rowcolor[HTML]{EAFAF1} 
\textbf{UniComp} & 64f/10\% & 56.2 & 65.5 & 256f/25\% & \textbf{58.8} & \textbf{71.0} \\
\midrule
VisionZip & 64f/25\% & 56.3 & 65.6 & 320f/20\% & 56.6 & 70.2 \\
HoliTom & 64f/25\% & 57.5 & 66.6 & 320f/20\% & \textit{OOM} & \textit{OOM} \\
\rowcolor[HTML]{EAFAF1} 
\textbf{UniComp} & 64f/25\% & \textbf{58.0} & \textbf{66.9} & 320f/20\% & \textbf{57.9} & \textbf{70.3} \\
\bottomrule
\end{tabular}
\label{tab:llava_vid}
\end{table}


\begin{table}[t]
\centering
\small
\setlength{\tabcolsep}{3pt}
\renewcommand{\arraystretch}{0.9}
\caption{Ablation study of Frame Group Fusion (FGF) and Spatial Dynamic Compression (SDC) with Token Allocation (TA) on multiple benchmarks under 32 and 320 frames setting on LLaVA-OneVision-7B with 20\% retained ratio. The baseline of 320-frame is bad since it exceeds the max token limitation, and the compression need to do is understand more frames under a token limitation.}
\begin{tabular}{lc|cccc}
\toprule
\textbf{Method} & \textbf{Frame/Retain} & \textbf{LVB} & \textbf{Ego} & \textbf{MLVU} & \textbf{VMME} \\
\midrule
\rowcolor{gray!15}
Vanilla & 32f / 100\% & 56.3 & 60.4 & \textbf{64.7} & 58.4 \\
+FGF & 32f / 92\% & 56.8 & 60.4 & \textbf{64.7} & \textbf{58.8} \\
\rowcolor[HTML]{EAFAF1} 
\textbf{+FGF+SAC$_{\text{TA}}$} & 32f / 20\% & \textbf{57.7} & \textbf{61.1} & 64.4 & 58.7 \\
\midrule
\rowcolor{gray!15}
Vanilla & 320f / 100\% & 53.3 & 59.3 & 61.1 & 56.5 \\
+FGF &320f / 61\% & 56.1 & 58.4 & 62.6 & 59 \\
\rowcolor[HTML]{EAFAF1} 
\textbf{+FGF+SAC$_{\text{TA}}$} & 320f / 10\% & \textbf{58.5} & \textbf{61.2}& \textbf{69.4} & \textbf{60.7} \\
\bottomrule
\end{tabular}
\label{tab:ablation}
\end{table}

\section{Experiments}

\subsection{Experiments Settings}

\paragraph{Benchmarks.} We evaluate \textit{UniComp} on four video understanding benchmarks featuring long videos: LongVideoBench~\cite{wu2024longvideobench}, EgoSchema~\cite{mangalam2023egoschema}, MLVU~\cite{zhou2024mlvu}, and VideoMME~\cite{fu2025video}.
These benchmarks cover a wide spectrum of video comprehension scenarios, including temporal reasoning, causal understanding, and multi-step event inference. Notably, they contains long videos with varying levels of semantic complexity and temporal dynamics, making them ideal for testing the ability of compression methods to preserve key visual information under limited computational budgets. 

\noindent
\textbf{Implementation Details.} 
We evaluate \textit{UniComp} on LLaVA-OneVision-7B~\cite{li2024llava}, LLaVA-Video-7B~\cite{zhang2024video} and Eagle2.5~\cite{chen2025eagle}. 
LLaVA-OneVision-7B~\cite{li2024llava} is a general-purpose video-language model, widely adopted by existing video compression and token selection studies. In contrast, Eagle2.5~\cite{chen2025eagle} represents the state-of-the-art long-video understanding model, capable of processing hundreds of frames as input. This allows us to validate \textit{UniComp}’s scalability and effectiveness in handling hour-long videos that demand both temporal coherence and information selectivity. All experiments evaluated with Lmms-Eval~\cite{zhang2025lmms} on H800 GPUs. \textit{UniComp} only have two hyper-parameters, and for all experiments, $U_f$ is set to 0.005, $U_c$ is set to 0.2.

\noindent
\textbf{Compared Methods.}
We conduct a comprehensive comparison between our proposed \textit{UniComp} and several strong training-free compression baselines.
Specifically, we include six representative approaches widely adopted in recent video-language compression research. FastV~\cite{chen2024image}, which selects key visual tokens during the prefilling stage based on attention correlations; PDrop~\cite{xing2024pyramiddrop}, which prunes vision tokens within partitioned LLM stages guided by both image and instruction tokens; DyCoke~\cite{tao2025dycoke}, which applies temporal merging prior to LLM input and dynamic KV cache pruning during decoding; PruneVid~\cite{huang2024prunevid}, which eliminates video redundancy through spatio-temporal token clustering. 
Moreover, we systematically compare three SOTA methods under different compression ratios and input frames - VisionZip~\cite{yang2025visionzip}, FastVID~\cite{shen2025fastvid} and HoliTom~\cite{shao2025holitom} - using their official codes and recommended hyper-parameters.

\subsection{Main Results}
\paragraph{Comparisons with State-of-the-Art Methods.}
As shown in Tab.~\ref{tab:results1}, we compare \textbf{\textit{UniComp}} with representative video compression methods at frame-retention ratios of 25\%, 20\%, 15\%, and 10\%, with accuracy averaged across benchmarks.

\textbf{\textit{UniComp}} consistently achieves the highest average accuracy under all settings. At 25\% retention, it reaches 60.78\%, outperforming FastV by 2.18 points; at 20\%, it achieves 60.48\%, exceeding VisionZip by 1.2 points; and at 10\% retention, it attains 59.80\%, surpassing HoliTom by 0.9 points. Overall, \textit{UniComp} maintains robust accuracy under extreme compression, enabling fast Video LLM inference without loss of essential visual information.



\paragraph{Performance Comparison Across Different Backbones.}
On Tab.~\ref{tab:llava_vid}, we compared \textit{UniComp} with SOTA methods on LLaVA-Video-7B. Additionally, in Supplementary, we present results on Eagle2.5, which is designed for long video understanding and adopts a distinct model architecture. The hyper-parameters of \textit{UniComp} are shared with all models, which shows the robustness and transferability of \textit{UniComp}.


\paragraph{Performance Scaling with more frames.}
To evaluate scalability, we increase the input from 32 frames to 320 frames and compress to a maximum of 6,272 tokens, comparing \textbf{\textit{UniComp}} with other methods across multiple benchmarks on LLaVA-OneVision-7B.

As shown in Tab.~\ref{tab:compression_results2}, \textbf{\textit{UniComp}} consistently achieves the highest average accuracy under the same token retention. Compared to the next-best, it achieves 62.4\% on EgoSchema at 128-frame input (a 0.9-point improvement), 57.4 on LongVideoBench at 256-frame input (a 1.2-point improvement), and 69.4\% on MLVU at 320-frame input (also a 1.2 points improvement). At 320-frame input setting, the overall average accuracy is 62.45\%, outperforming the next-best method by 1.02\%.

Compared to the uncompressed baseline, our approach improves average accuracy by 2.5\% under the same token budget at the 320-frame input setting. These consistent gains demonstrate the robustness of \textit{UniComp} for long video inputs. Furthermore, at equivalent computational efficiency, performance can be further improved by increasing the number of input frames and leveraging compression.



\begin{figure}
\centering
\includegraphics[width=0.43\textwidth]{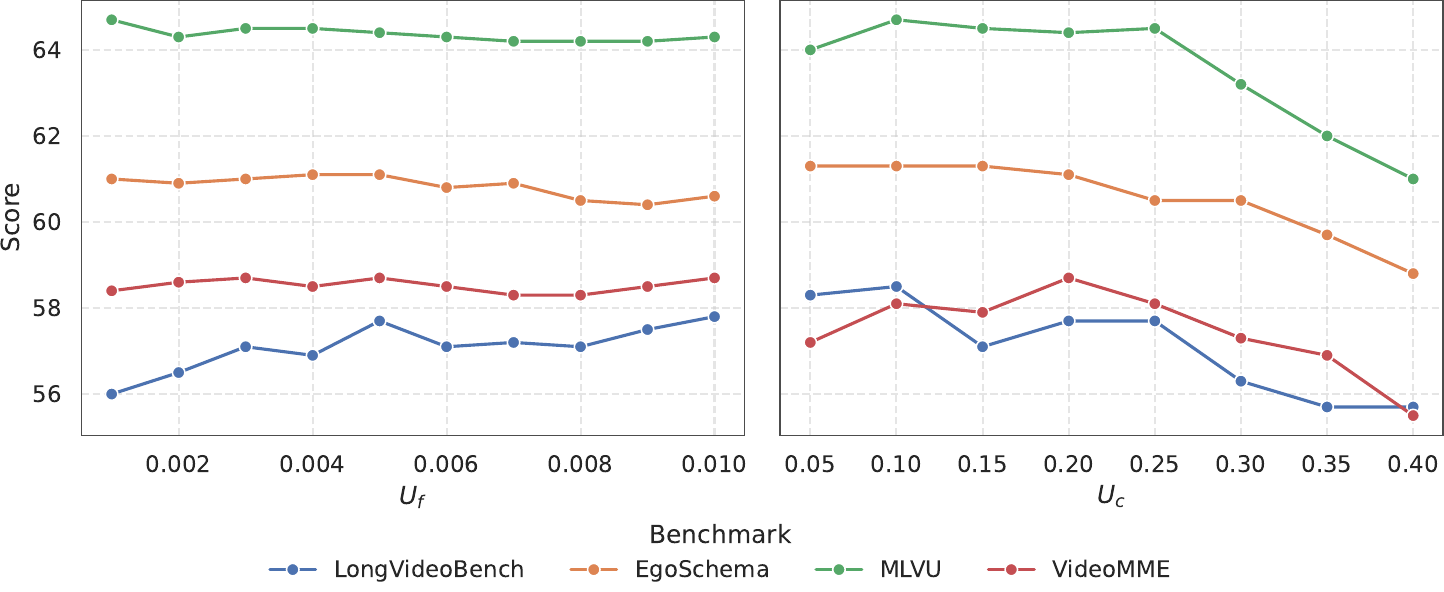}
\caption{Hyper-parameters analysis across four benchmarks. \textbf{Left:} Analysis of $U_f$ in FGF, it demonstrates the stability of the performance impact. \textbf{Right:} Analysis of $U_c$ in SDC, even though the current setting of 0.2 has already achieved excellent performance, further improvements can be achieved in these benchmarks through parameter fine-tuning.}
\label{fig:param}
\end{figure}

\paragraph{Parameters Analysis.}
We further analyze the impact of two key hyper-parameters across four benchmarks, as shown in Fig.~\ref{fig:param}.
The frame-level uniqueness threshold $U_f$ in FGF (left) shows stable performance across a wide range of values, indicating the robustness of our temporal grouping strategy.
For the token-level uniqueness threshold $U_c$ in SDC (right), our default setting ($U_c=0.2$) already achieves strong performance, while slight tuning can yield additional gains. These results suggest that \textit{UniComp} effectively balances information preservation and redundancy reduction.

\begin{figure}
\centering
\begin{minipage}[b]{0.49\linewidth}
    \centering
    \includegraphics[width=\linewidth]{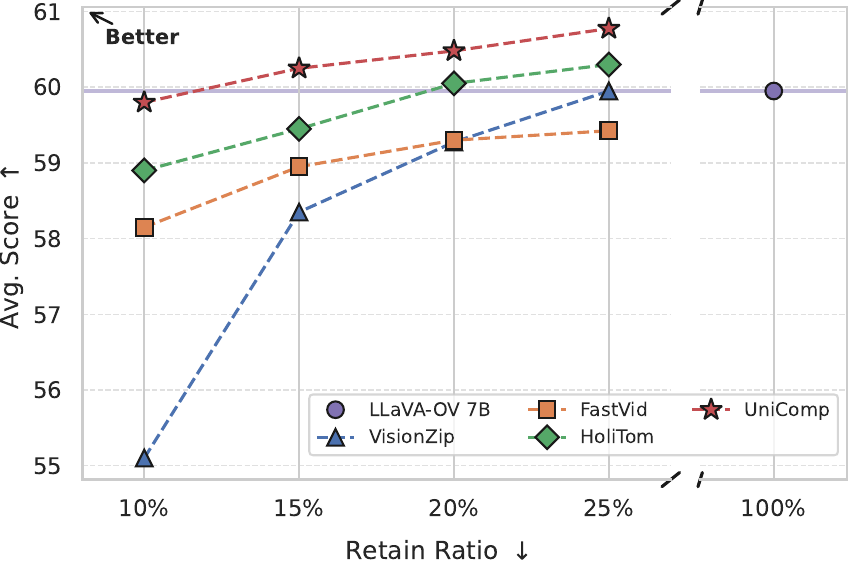}
\end{minipage}\hfill
\begin{minipage}[b]{0.49\linewidth}
    \centering
    \includegraphics[width=\linewidth]{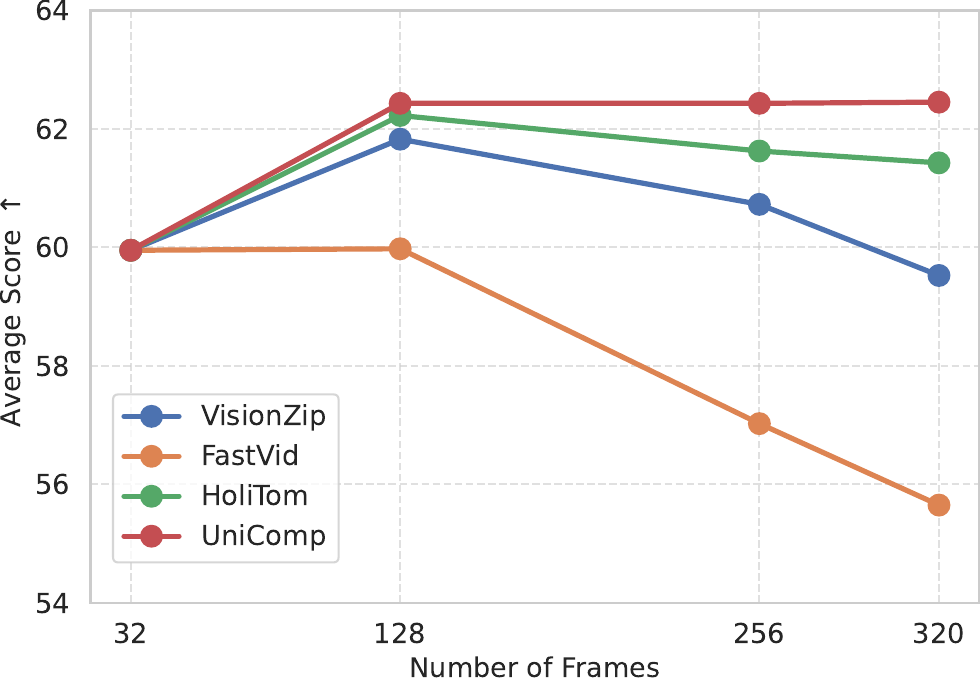}
\end{minipage}
\caption{\textbf{Left:} comparison with SOTA compression methods under different retained ratios, \textit{UniComp} outperforms all and even better than full tokens without compression. \textbf{Right: } comparison with SOTA compression methods under different frames (but same token limitation 32 frames $\times$ 196 tokens)}
\label{fig:efficient}
\end{figure}

\subsection{Efficiency Comparison}

\paragraph{Overall Comparison.}
Fig.~\ref{fig:efficient} (left) illustrates the performance under different retained ratios. \textit{UniComp} consistently surpasses all SOTA training-free compression methods (FastVid, VisionZip, and HoliTom) and even exceeds the uncompressed full-token baseline. This demonstrates that our uniqueness-driven compression not only preserves but enhances representation efficiency, effectively filtering redundant tokens while retaining highly informative ones.

\begin{figure}
\centering
\includegraphics[width=0.43\textwidth]{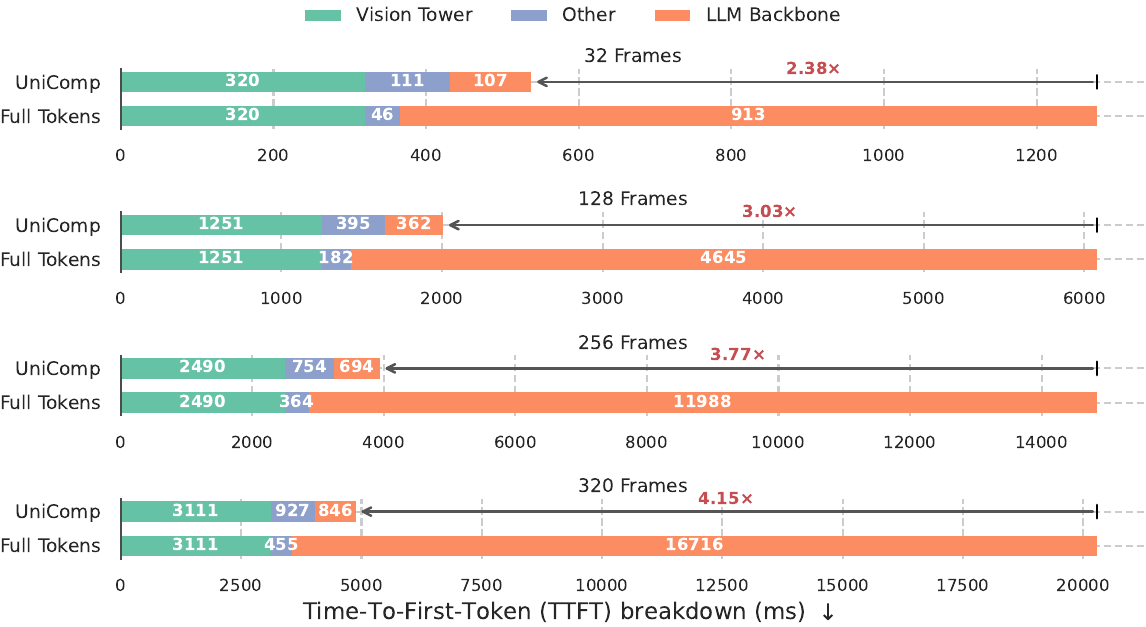}
\caption{Comparison with Vanilla method with full set token on efficiency. \textit{UniComp} achieves up to 4× faster Time-To-First-Token (TTFT), demonstrating superior efficiency on long videos.}
\label{fig:ttft-tp}
\end{figure}

\paragraph{Scalability with Frame Length.}
As shown in Fig.~\ref{fig:efficient} (right), \textit{UniComp} maintains superior performance as the input length scales up from 32 to 320 frames under the same token limit (32 × 196). Unlike FastVid and VisionZip, whose accuracy rapidly declines due to excessive temporal redundancy, \textit{UniComp} and HoliTom both remain stable, while \textit{UniComp} achieves the best trade-off between temporal coverage and token efficiency.

\paragraph{Efficiency Analysis.}
Fig.~\ref{fig:ttft-tp} compares computational efficiency. \textit{UniComp} achieves a 4.15× reduction in \textit{Time-To-First-Token (TTFT)} with full-token inference under 320 frames with 10\% retained ratio. \textit{UniComp} provides a balanced gain in both performance and efficiency, making it a practical compression solution.

\subsection{Ablation Study}
\paragraph{Ablation on different modules with different frames}
We conduct a comprehensive ablation study to evaluate the effectiveness of each module in UniComp under both 32-frame and 320-frame settings (Tab.~\ref{tab:ablation}).
In the 32-frame setting, introducing the Frame Group Fusion (FGF) module already improves performance with an automatic 92\% token retention ratio. Further integrating Spatial Dynamic Compression (SDC) with Token Allocation (TA) — retaining only 20\% of the total tokens — yields additional gains on LongVideoBench and EgoSchema, while maintaining strong performance on MLVU and VideoMME. 

To better show the ability of Frame Group Fusion, we evaluate under large frames input (320-frame) setting, the FGF module removes 39\% of redundant frames while sustaining high performance. While the vanilla model suffers from token overload, \textit{UniComp} effectively compresses redundant frames and retains semantic information, confirming its scalability to long-video scenarios. 


\paragraph{Ablation on individual module designs.}
Tab.~\ref{tab:ablation_detail} presents detailed results analyzing each module’s internal design.
For FGF, fusing semantically similar frames outperforms using only the first frame, confirming the necessity of temporal fusion.
For Token Allocation (TA), our adaptive strategy surpasses uniform allocation, highlighting the benefit of assigning more tokens to semantically unique frames.
For Spatial Dynamic Compression (SDC), the first two rows show that uniqueness-guided token selection outperforms attention-based selection. Comparing the third row with the second, our SDC further surpasses simply selecting the top-k tokens. Moreover, variants that disable token fusion or use the selection order without restoring the original token IDs consistently lead to performance drops, highlighting the effectiveness of our full pipeline.





\begin{table}[t]
\centering
\small
\setlength{\tabcolsep}{3pt}
\renewcommand{\arraystretch}{0.75}
\caption{Ablation study of different modules (FGF, TA, SDC) under 32-frame setting with 20\% retained ratio. \textbf{`first frame'}: only keep the first frame during FGF. \textbf{`Uniform'}: uniform token allocation w/o TA. \textbf{`attn topk'}: keep top-k attention based tokens like VisionZip~\cite{yang2025visionzip}. \textbf{`unique topk'}: keep top-k uniqueness based tokens. 
\textbf{`-fusion'}: not fuse neighbor tokens. \textbf{`-ids order'}: tokens input LLM with uniqueness order instead of ids order.}
\begin{tabular}{lcccc}
\toprule
\textbf{Method} & \textbf{LVB} & \textbf{Ego} & \textbf{MLVU} & \textbf{VMME} \\
\midrule
\rowcolor{gray!15}
\textbf{Baseline (32f)} & 56.3 & 60.4 & 64.7 & 58.4 \\
\midrule
\textbf{FGF:} & & & & \\
\quad first frame & 56.9 & 60.8 & 64.2 & 58.5 \\
\rowcolor[HTML]{EAFAF1} 
\quad \textbf{fusion (ours)} & \textbf{57.7} & \textbf{61.1} & \textbf{64.4} & \textbf{58.7} \\
\midrule
\textbf{TA:} & & & & \\
\quad Uniform & 57.3 & 60.5 & 61.4 & 56.4 \\
\rowcolor[HTML]{EAFAF1} 
\quad \textbf{Allocation (ours)} & \textbf{57.7} & \textbf{61.1} & \textbf{64.4} & \textbf{58.7} \\
\midrule
\textbf{SDC:} & & & & \\
\quad attn topk & 56.4 & 59.4 & 64.0 & 57.4 \\
\quad unique topk & 56.7 & 60.0 & 64.3 & 57.3 \\
\rowcolor[HTML]{EAFAF1} 
\quad \textbf{SDC (our)} & \textbf{57.7} & \textbf{61.1} & \textbf{64.4} & \textbf{58.7} \\
\quad - fusion & 57.2 & 60.8 & 63.9 & 58.0 \\
\quad - ids order & 57.0 & 61.1 & 64.7 & 58.4 \\
\bottomrule
\end{tabular}
\label{tab:ablation_detail}
\end{table}

\section{Conclusion}

In this work, we proposed a new paradigm for video token compression by prioritizing information uniqueness instead of attention. Grounded in an information-theoretic formulation, we developed UniComp, a lightweight framework that efficiently reduces redundancy across temporal and spatial dimensions. UniComp requires only two hyper-parameters and minimal code changes for deployment. Extensive experiments demonstrate that UniComp consistently outperforms existing methods, achieving superior semantic fidelity and efficiency, and offering a practical and powerful solution for scaling up multimodal models.

\maketitlesupplementary

\section{Derivation of the Reconstruction Error Bound}
\label{sec:AppendixA}

In this section, we provide the algebraic derivation supporting the upper bound relation
\begin{equation}
\mathcal{E}(\mathcal{S}) = \sum_{j\in\mathcal{X}}\|x_j - \hat x_j\|^2 \leq 2 \sum_{j \in \mathcal{X}} \min_{i\in\mathcal{S}} u_{ij}
\end{equation}
where $\hat x_j$ is the reconstruction of $x_j$ using tokens in $\mathcal{S}$. where $\hat x_j = \sum_{i\in\mathcal{S}} w_{ij} x_i$, $\sum_i w_{ij}=1$, $w_{ij}\ge0$, and $u_{ij} = 1-s_{ij}$, and $s_{ij}$ denotes the cosine similarity between normalized features. The weight $w_{ij}$ reflects the relative similarity between $x_i$ and $x_j$:
\begin{equation}
w_{ij} = \frac{\exp(s_{ij})}{\sum_{i\in\mathcal{S}} \exp(s_{ij})}
\end{equation}



\paragraph{Step 1. Expansion of the reconstruction error.}
Given normalized feature vectors $\|x_i\|=\|x_j\|=1$, we can expand the reconstruction error term as
\begin{align}
\|x_j - \hat x_j\|^2
&= \|x_j - \sum_i w_{ij}x_i\|^2 \\
&= \|x_j\|^2 - 2\sum_i w_{ij} x_j^\top x_i + \sum_{i,k} w_{ij}w_{kj} x_i^\top x_k \\
&= 1 - 2\sum_i w_{ij}s_{ij} + \sum_{i,k} w_{ij}w_{kj}s_{ik}
\label{eq:error_expansion}
\end{align}

\paragraph{Step 2. Bounding the cross-term.}
Since all pairwise similarities satisfy $s_{ik}\le1$, the last term in Eq.~\eqref{eq:error_expansion} can be upper-bounded by
\begin{equation}
\sum_{i,k} w_{ij}w_{kj}s_{ik} \le \sum_{i,k} w_{ij}w_{kj} = \Big(\sum_i w_{ij}\Big)^2 = 1
\end{equation}
Substituting back gives a coarse upper bound:
\begin{equation}
\|x_j - \hat x_j\|^2 \le 2(1 - \sum_i w_{ij}s_{ij}) \label{eq:upperbound1}
\end{equation}

\paragraph{Step 3. Approximating with the most similar token.}


Since all the discarded tokens has a very similar token has been selected, so, one selected token $x_{i^*}$ could dominate the reconstruction (i.e., $w_{i^*j}\approx1$ and $s_{i^*j}=\max_i s_{ij}$), then
\begin{equation}
\sum_i w_{ij}s_{ij} \approx s_{i^*j}
\end{equation}
and Eq.~\eqref{eq:upperbound1} becomes
\begin{equation}
\|x_j - \hat x_j\|^2 \leq 2(1 - s_{i^*j}) \label{eq:upperbound2}
\end{equation}
Thus, we obtain the final form:
\begin{equation}
\sum_{j\in\mathcal{X}}\|x_j - \hat x_j\|^2 \leq 2\sum_{j\in\mathcal{X}} \min_{i\in\mathcal{S}} (1 - s_{ij}) = 2 \sum_{j\in\mathcal{X}} \min_{i\in\mathcal{S}} u_{ij}
\label{eq:finalbound}
\end{equation}

\paragraph{Step 4. Interpretation.}
This result implies that the reconstruction error of any discarded token $x_j$ is upper-bounded by the angular distance to its most similar selected token in the normalized feature space.  
The bound holds under the convex-combination constraint ($w_{ij}\ge0$, $\sum_i w_{ij}=1$), which ensures that $\hat x_j$ lies inside the convex hull of the selected features.  
Thus, a higher pairwise similarity $s_{ij}$ directly corresponds to a smaller reconstruction uncertainty, providing a principled link between similarity-based redundancy reduction and information preservation.

\section{Auto Compression Analysis}

\paragraph{More details of main experiment settings.} 

In the main experiments, Frame Group Fusion (FGF) may directly compress video lower than a given retained ratio due to temporal correlations and redundancies, so that Spatial Dynamic Compression (SDC) may be bypassed when the resulting token count is below the target. Without a preset ratio, the model performs automatic compression based on both temporal and spatial redundancies.

\begin{figure}
\centering
\includegraphics[width=0.5\textwidth]{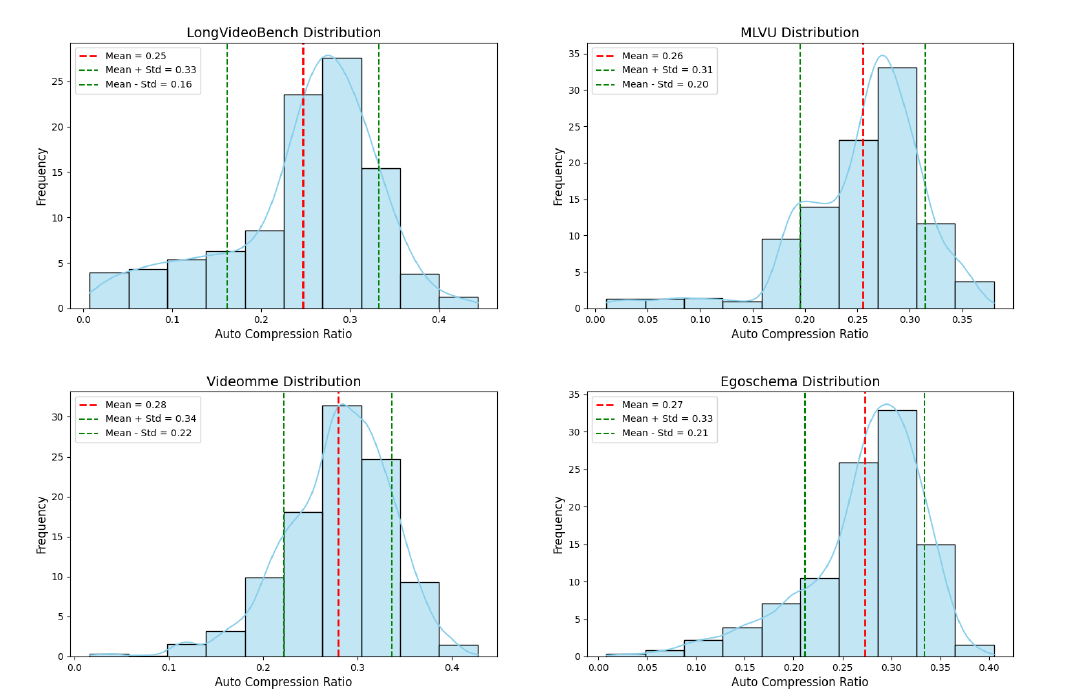}
\caption{Auto compression on different benchmarks of LLava-OneVision-7B. It demonstrates the information redundancy of videos in different benchmarks.}
\label{fig:auto_distribution}
\end{figure}

\paragraph{Fully automatically compression without retention limitation.}
Since \textbf{\textit{UniComp}} supports fully automatic video compression (results shown in Table~\ref{tab:auto}), we visualize the retention tokens under the automatic setting across four benchmarks in Figure~\ref{fig:auto_distribution}. The visualization also shows the information redundancy of videos in different benchmarks, which means a lot of videos can be represented with only a few tokens, and LongVideoBench exhibits the highest redundancy.

\begin{figure}
\centering
\includegraphics[width=0.4\textwidth]{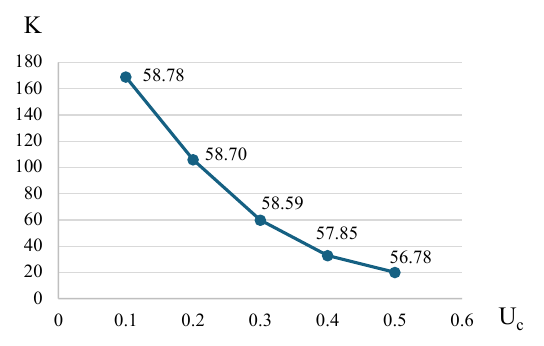}
\caption{Each frame average retained token number $K$ under auto compression settings with $U_c$ on VideoMME with LLava-OneVision-7B, which shows the best retained tokens $K$ of different $U_c$, given a detailed understanding of hyper-parameter $U_c$.}
\label{fig:uk}
\end{figure}

Moreover, in Figure~\ref{fig:uk}, we show the average automatically compressed tokens on VideoMME, which has equal numbers of short, medium, and long videos (100 each). It shows the automatically compressed tokens, the hyper-parameter $U_c$, and their performance on VideoMME.

\begin{table}[t]
\centering
\small
\setlength{\tabcolsep}{3pt}
\renewcommand{\arraystretch}{0.8}
\caption{\textit{UniComp} fully automatically compress results on LLava-OneVision-7B. It may be slightly lower than the main experiments since this experiment will not limit the retention ratio and it could compress much lower (e.g. 2\% as shown in Figure~\ref{fig:auto_distribution}).}
\begin{tabular}{lccccc}
\toprule
\textbf{Method} & \textbf{Retention Ratio} & \textbf{LVB} & \textbf{Ego} & \textbf{MLVU} & \textbf{VMME} \\
\midrule
\rowcolor{gray!15}
Baseline & 100\% & 56.3 & 60.4 & 64.7 & 58.4 \\
\textbf{UniComp (auto)} & 26.3\% & 57.3 & 61.5 & 64.0 & 58.8 \\
\bottomrule
\end{tabular}
\label{tab:auto}
\end{table}

\begin{table}[t]
\centering
\small
\setlength{\tabcolsep}{4pt}
\renewcommand{\arraystretch}{1}
\caption{Comparison on sub-tasks that may be heavily affected by token compression on VideoMME. Temporal Perception (TP), Object Recognition (OR), Action Recognition (AR), and Counting Problem (CP).}
\begin{tabular}{lcccc}
\toprule
\textbf{Method} & \textbf{TP} & \textbf{OR} & \textbf{AR} & \textbf{CP} \\
\midrule
\rowcolor{gray!15}
Base (32f) & 63.6 & 65.3 & 54.3 & 37.3 \\
VisionZip & 61.8 & 64.4 & 58.1 & 35.4 \\
FastVid   & 61.8 & 63.8 & 54.0 & 35.4 \\
HoliTom   & 60.0 & 64.1 & 55.3 & 37.3 \\
\rowcolor[HTML]{EAFAF1} 
UniComp   & \textbf{63.6} & \textbf{65.5} & \textbf{58.5} & \textbf{39.6} \\
\bottomrule
\end{tabular}
\label{tab:subtask}
\end{table}

\begin{table}[t]
\centering
\small
\setlength{\tabcolsep}{3pt}
\renewcommand{\arraystretch}{0.8}
\caption{Performance comparison on VideoMME under long video settings on \textbf{Eagle-2.5 which is designed for hour long video} with \textbf{hundreds of frames input} to show the real video understanding ability. Each experiment compresses to 64 frames' tokens (64*256). HoliTom* means without the inner-LLM mode (w/o M), as this mode is very difficult to implement in other models.}
\begin{tabular}{lc|cccc}
\toprule
\textbf{Method} & \textbf{Frames/Retain} & \textbf{All} & \textbf{Short} & \textbf{Medium} & \textbf{Long} \\
\midrule
\rowcolor{gray!15}
Vanilla & 64f / 100\% & 68.9 & 80.6 & 67.1 & 59.1 \\
\midrule
VisionZip & 128f / 50\% & 67.8 & 78.9 & 65.1 & 59.4 \\
HoliTom*  & 128f / 50\%  & 68.4 & 81.0 & 65.4 & 58.9 \\
\rowcolor[HTML]{EAFAF1} 
\textbf{UniComp} & 128f / 50\%  & \textbf{70.1} & \textbf{81.4} & \textbf{69.0} & \textbf{59.8} \\
\midrule
VisionZip & 256f / 25\%  & 66.3 & 76.0 & 64.0 & 58.9 \\
HoliTom*  & 256f / 25\%  & 67.9 & 78.7 & 65.4 & 59.4 \\
\rowcolor[HTML]{EAFAF1} 
\textbf{UniComp} & 256f / 25\%  & \textbf{70.4} & \textbf{80.8} & \textbf{69.7} & \textbf{60.8} \\
\midrule
VisionZip & 512f / 12.5\% & 65.0 & 72.9 & 62.9 & 59.2 \\
HoliTom* & 512f / 12.5\% & 67.4 & 77.9 & 65.4 & 58.9 \\
\rowcolor[HTML]{EAFAF1} 
\textbf{UniComp} & 512f / 12.5\% & \textbf{70.7} & \textbf{80.3} & \textbf{69.3} & \textbf{62.4} \\
\bottomrule
\end{tabular}
\label{tab:eagle}
\end{table}

\section{More Experiments}
We further evaluate \textit{UniComp} on sensitive sub-tasks and long-video scenarios. As shown in Table~\ref{tab:subtask}, \textit{UniComp} achieves the best performance across all four sub-tasks (TP, OR, AR, CP). Notably, it preserves temporal and object information as well as the full-frame baseline while offering clear gains on action recognition and counting, indicating its stronger ability in retaining critical visual cues during compression.

For long-video evaluation on Eagle-2.5 (Table~\ref{tab:eagle}), which is naturally designed for hour-long videos understanding, \textit{UniComp} consistently outperforms VisionZip and HoliTom under all compression settings, from 128f to 512f inputs compressed to 64-frame tokens. 

\section{ViT keys versus other representations in SDC}
Using ViT Keys/Values or Last layer feats as representations to calculate uniqueness. See Table \ref{keys}.
\begin{table}[h]
\footnotesize
\centering
\caption{Comparison of ViT keys versus other representations}
\begin{tabular}{lcccc}
\hline
  & Longvideo & Egoschema & MLVU & Videomme \\
\hline
Attn Values          & 57.4      & 58.8  & 63.6 & 57.4  \\
Last layer feats   & 55.8      &\textbf{61.1}      & \textbf{64.7} & 58.4     \\
\textbf{Attn Keys (ours) } & \textbf{57.7}      & \textbf{61.1}     & 64.4 & \textbf{58.7}     \\
\hline
\end{tabular}
\vspace{-10pt}
\label{keys}
\end{table}

\section{Breakdown TTFT latency}

Tab.~\ref{tab:ttft_breakdown} breaks down TTFT latency, showing that SDC contributes most to the overhead, while the ``Other'' proportion increases with the number of frames. Tab.~\ref{tab:method_comparison} compares methods under the 320-frame setting. Our algorithm introduces less overhead than others.

\begin{table}[h]
    \centering
\small
    \setlength{\tabcolsep}{4pt}
    \caption{TTFT breakdown (ms) under the 320-frame setting with 10\% retained ratio..}
    \begin{tabular}{lccccc}
        \hline
        \textbf{Frame} & \textbf{Compress} & \textbf{FGF} & \textbf{TA} & \textbf{SDC} & \textbf{Other} \\
        \hline
        32  & 111 & 1.60 & 1.08 & 61.78 & 46.54 \\
        128 & 395 & 6.21 & 0.93 & 206.20 & 181.66 \\
        256 & 754 & 13.48 & 1.41 & 367.82 & 371.30 \\
        320 & 927 & 17.73 & 1.73 & 447.38 & 460.15 \\
        \hline
    \end{tabular}
    \label{tab:ttft_breakdown}
\end{table}

\begin{table}[h]
    \centering
\small
    \setlength{\tabcolsep}{6pt}
    \caption{Comparison of different methods under the 320-frame setting with 10\% retained ratio.}
    \begin{tabular}{lcccc}
        \hline
        \textbf{Method} & \textbf{ViT} & \textbf{Compress} & \textbf{LLM} & \textbf{Avg. Score} \\
        \hline
        VisionZip & 3111 & 474  & 948   & 59.53 \\
        HoliTom   & 3111 & 2020 & 993   & 61.43 \\
        \textbf{UniComp} & 3111 & 927  & 846   & \textbf{62.45} \\
        Full Tokens & 3111 & 455  & 16716 & 57.55 \\
        \hline
    \end{tabular}
    \label{tab:method_comparison}
\end{table}

\section{Comparisons on temporal grouping strategies}
Tab.\ref{fgf} shows comparison with temporal grouping strategies. Unlike standard attention methods that may overlook subtle changes, UniComp maximizes information retention. We preserve motion cues via weighted fusion (FGF), specifically anchored on the first frame to maintain trajectory boundaries (avoiding ambiguity in directional motion), and adaptive token allocation (TA) for unique movements. 
\begin{table}[htbp]
\footnotesize
\centering
\caption{Comparison on different temporal grouping strategies}
\begin{tabular}{lcccc}
\hline
\textbf{Method} & \textbf{LVB} & \textbf{Ego} & \textbf{MLVU} & \textbf{VMME} \\
\hline
Dyseg (FastVid)    & 51.6 & 56.5 & 58.8 & 53.7 \\
Redundancy-Aware (HoliTom)  & 57.0 & 60.1 & 61.6 & 55.6 \\
FGF (Ours)     & \textbf{57.7} & \textbf{61.1} & \textbf{64.4} & \textbf{58.7} \\
\hline
\end{tabular}
\label{fgf}
\end{table}

\section{Implementation Details}

Our method is implemented on the LLaVA-OneVision-7B, LLaVA-Video-7B and Eagle2.5-7B models. We conduct evaluation on NVIDIA H800 (80 GB) GPUs, while inference is tested on NVIDIA H20-3e (141 GB) GPUs to better reflect practical deployment scenarios. All benchmark evaluations are performed using LMMs-Eval.

\section{Qualitative Examples}

\subsection{Real example visualizations and workflow}
To further illustrate how \textit{UniComp}'s real behaves under different temporal dynamics, we present two real examples that visualize both the retained tokens and the detailed fusion patterns generated by our pipeline.

\begin{figure*}
\centering
\includegraphics[width=\textwidth]{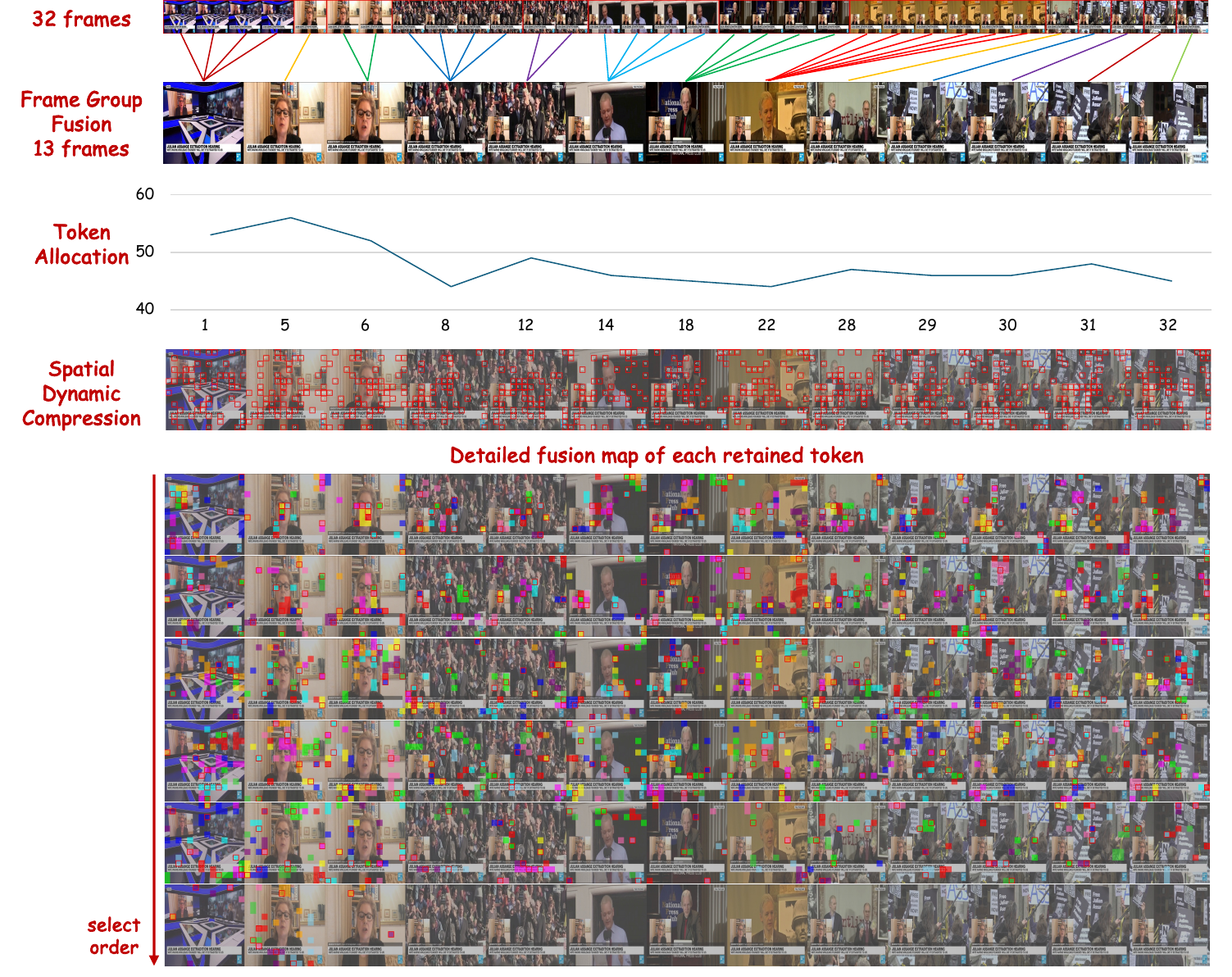}
\caption{\textbf{A real example that scenes switch frequently} of \textit{UniComp} on LLaVA-OneVision-7B with 32 frames input under 10\% retention ratio. The images are \textbf{high-fidelity} and can be \textbf{zoomed in to view details}. After Spatial Dynamic Compression, it shows the retained tokens with the selection order labeled white. Each line of detailed fusion map shows 10 tokens, same color token will be fused to the token with red rectangle. The order from top to bottom is the selection order, which means higher one is more unique.}
\label{fig:flow_demo1}
\end{figure*}

Figure~\ref{fig:flow_demo1} shows a case with \textbf{frequent scene switches}. In such highly dynamic scenarios, consecutive frames exhibit strong semantic differences. \textit{UniComp} adaptively identifies these transitions through Frame Group Fusion (FGF), generating finer-grained groups. Token Allocation (TA) assigns more tokens to these semantically distinctive segments to preserve critical visual cues. During Spatial Dynamic Compression (SDC), tokens are selected based on their local uniqueness, with the selection order indicated by white labels. Tokens appearing earlier in the list exhibit higher uniqueness. In the fusion maps, tokens sharing the same color are fused into the representative token marked by a red rectangle. Despite significant scene changes, \textit{UniComp} consistently preserves the most informative regions, producing high-fidelity visual reconstructions that remain clear even when zoomed in.

\begin{figure*}
\centering
\includegraphics[width=\textwidth,height=0.8\textheight]{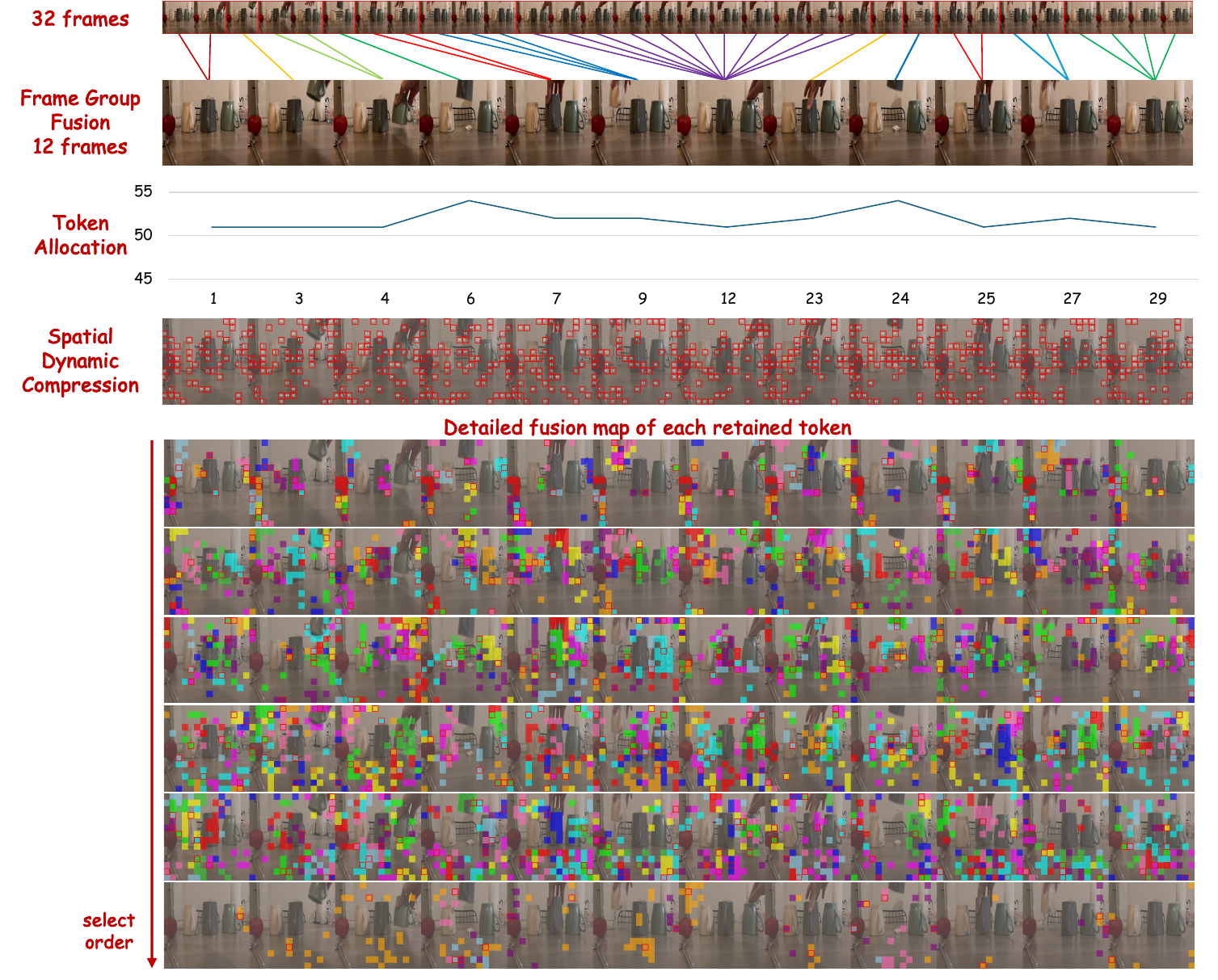}
\caption{\textbf{A real example that the scene remains nearly unchanged} of \textit{UniComp} on LLaVA-onvision-7B with 32 frames input under 10\% retention ratio. The images are \textbf{high-fidelity} and can be \textbf{zoomed in to view details}. After Spatial Dynamic Compression, it shows the retained tokens with the selection order labeled white. Each line of detailed fusion map shows 10 tokens, same color token will be fused to the token with red rectangle. The order from top to bottom is the selection order, which means higher one is more unique.}
\label{fig:flow_demo2}
\end{figure*}

Figure~\ref{fig:flow_demo2} presents a scenario where the scene remains \textbf{nearly unchanged}. Here, the global uniqueness across frames is low, leading FGF to merge many consecutive frames into a single stable semantic segment. Consequently, TA allocates fewer tokens to this region. SDC further identifies large areas of spatial redundancy, allowing many tokens to be merged into a small set of representative ones. Even under aggressive compression, \textit{UniComp} maintains coherent visual content and produces high-quality images.

Together, these examples demonstrate the adaptive nature of \textit{UniComp}: it allocates richer capacity to segments with high semantic variation while aggressively compressing redundant regions, achieving efficient yet information-preserving visual token compression.

\subsection{Performance under different compression ratio}
\begin{figure*}
\centering
\includegraphics[width=\textwidth,height=0.8\textheight]{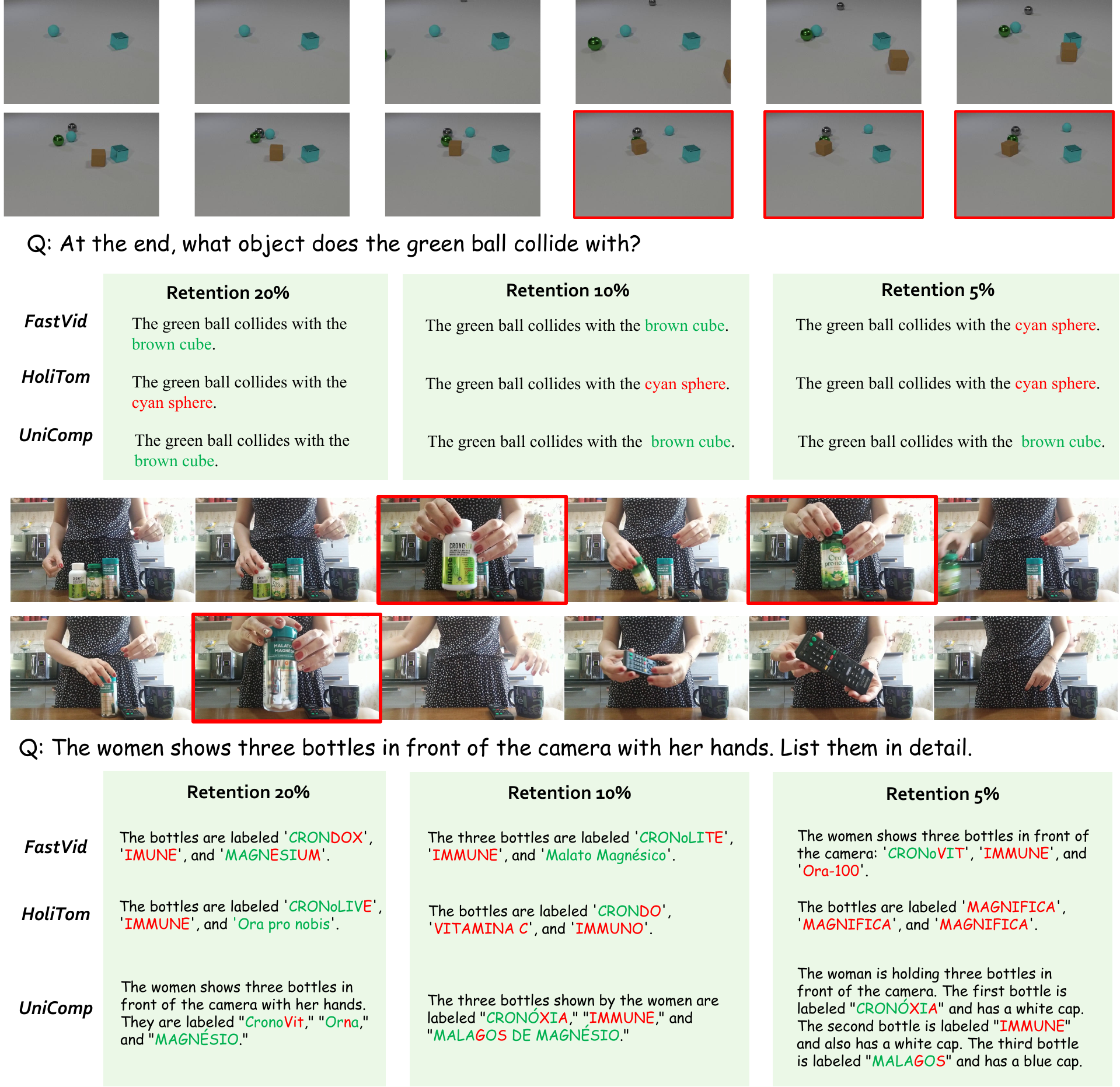}
\caption{Two examples of open-end question answer compare with FastVid and HoliTom.}
\label{fig:oe_demo1}
\end{figure*}

As shown in Figure~\ref{fig:oe_demo1}. We visualize two examples with open-end video question-answer under different compression ratio. It could be seen that \textit{UniComp} keep performing well.

\begin{figure*}
\centering
\includegraphics[width=\textwidth]{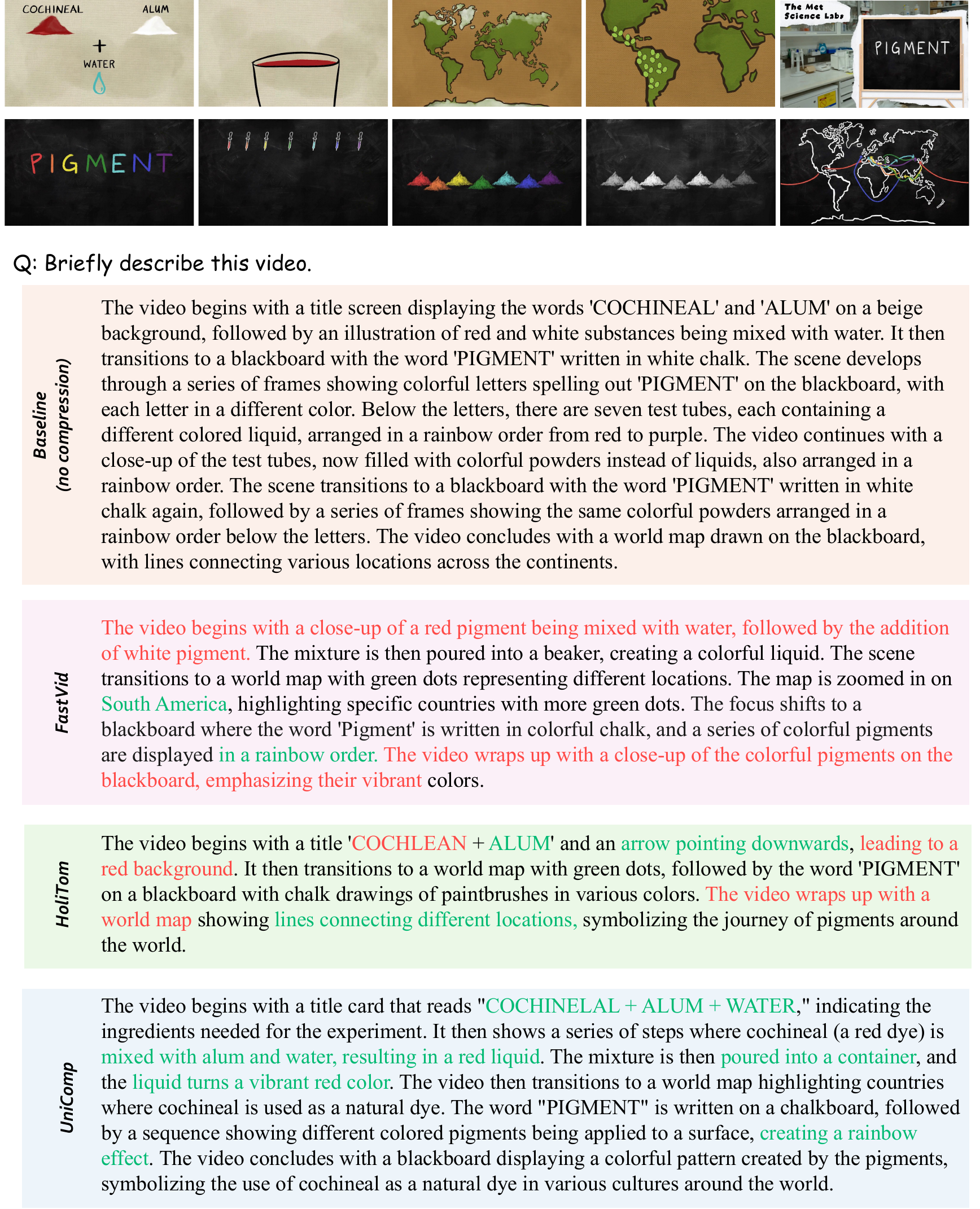}
\caption{An caption examples compare with FastVid and HoliTom.}
\label{fig:oe_demo2}
\end{figure*}

As shown in Figure~\ref{fig:oe_demo2}. We visualize an example with video caption generation task under 10\% compression ratio. \textit{UniComp} can still achieve strong understanding compared to other SOTA methods.

\subsection{Performance under different benchmarks}
\begin{figure*}
\centering
\includegraphics[width=\textwidth,height=0.8\textheight]{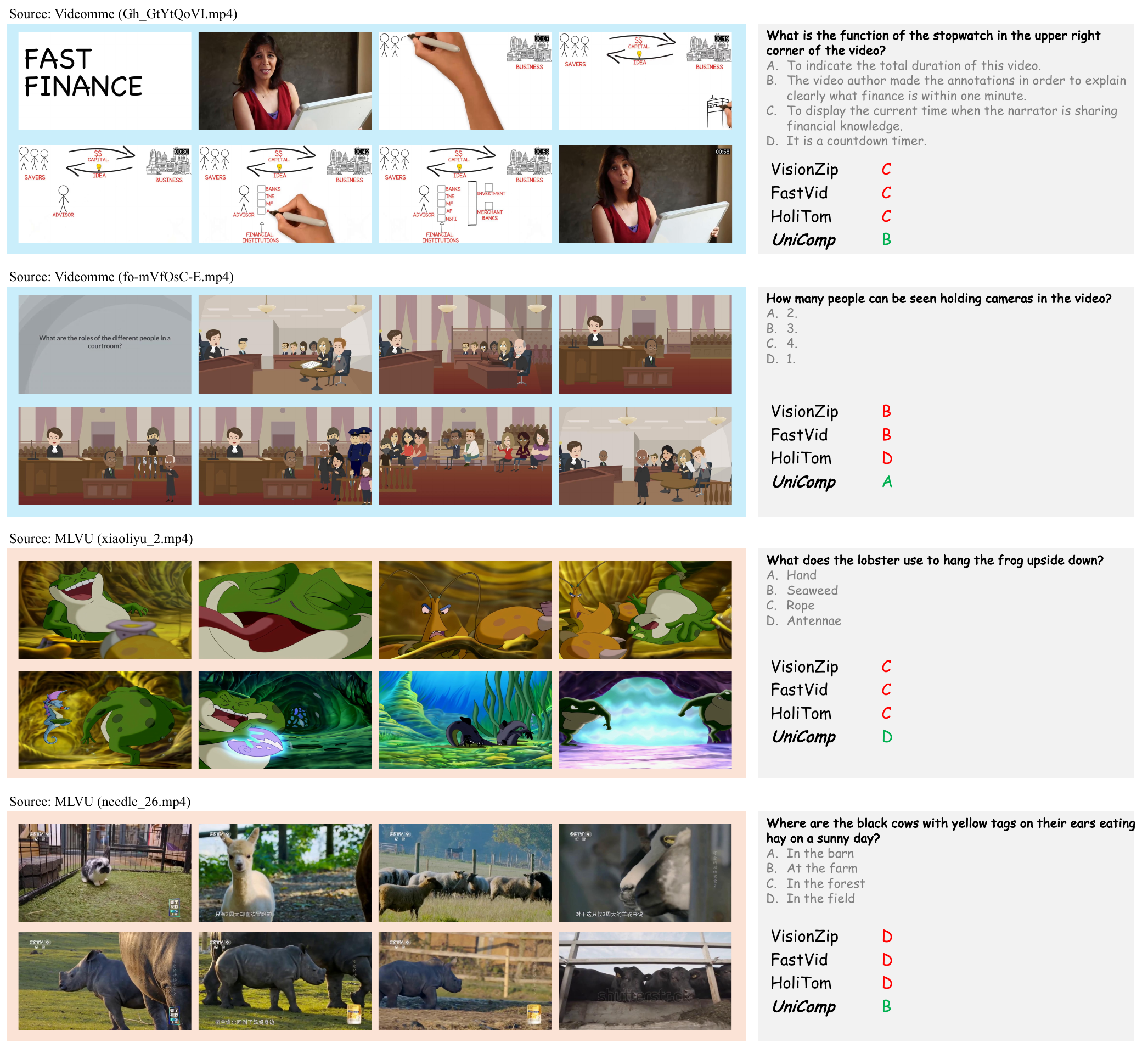}
\caption{Examples on VideoMME and MLVU.}
\label{fig:mcqa_demo1}
\end{figure*}

\begin{figure*}
\centering
\includegraphics[width=\textwidth,height=0.8\textheight]{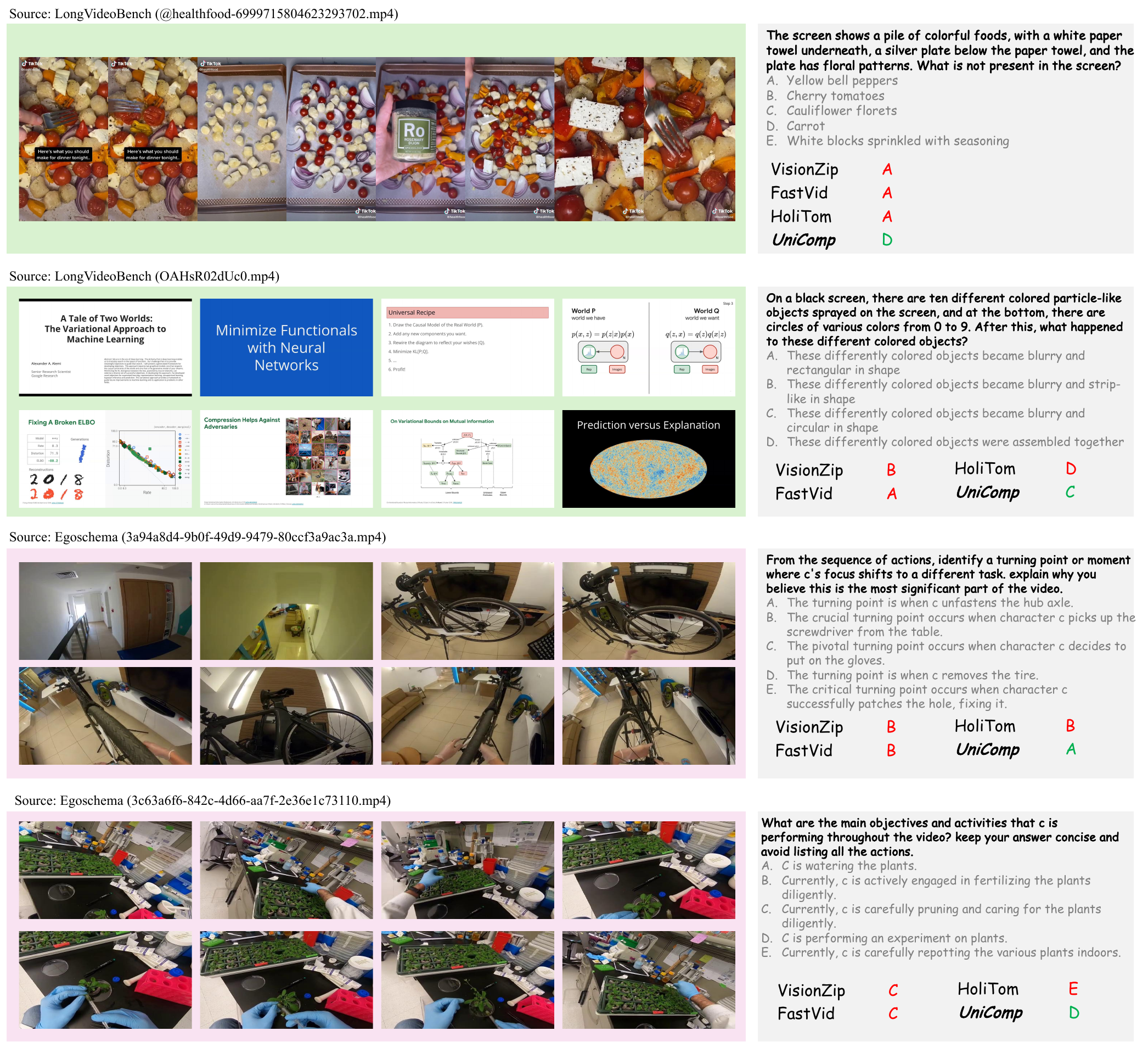}
\caption{Examples on LongVideoBench and Egoschema.}
\label{fig:mcqa_demo2}
\end{figure*}

As shown in Figure~\ref{fig:mcqa_demo1} and Figure~\ref{fig:mcqa_demo2}. We visualize examples on four benchmarks compared to SOTA compression methods.

    \clearpage

{
    \small
    \clearpage
    \bibliographystyle{ieeenat_fullname}
    \bibliography{main}
}
\end{document}